\pdfoutput=1

\documentclass[11pt]{article}

\usepackage[final]{acl}

\usepackage{times}
\usepackage{latexsym}
\usepackage{subcaption}
\usepackage{booktabs}
\usepackage{enumitem} 
\usepackage{siunitx}
\usepackage{caption}
\usepackage{amsmath} 
\usepackage{amssymb} 
\usepackage{cleveref}
\usepackage{tabularx}   

\usepackage[T1]{fontenc}

\usepackage[utf8]{inputenc}

\usepackage{microtype}

\usepackage{inconsolata}

\usepackage{graphicx}
\usepackage{float}
\usepackage[final,inline,nomargin,index]{fixme}  

\FXRegisterAuthor{tl}{atl}{\color{blue}Taesung}
\FXRegisterAuthor{hc}{ahc}{\color{red}Haw-Shiuan}  

\stepcounter{footnote} 

\title{Latent Traits and Cross-Task Transfer: \\ Deconstructing Dataset Interactions in LLM Fine-tuning}

\author{Shambhavi Krishna\textsuperscript{1,*} \quad 
        Atharva Naik\textsuperscript{1,*} \quad 
        Chaitali Agarwal\textsuperscript{1,*} \\
        \bf Sudharshan Govindan\textsuperscript{1,*} \quad 
        \bf Haw-Shiuan Chang\textsuperscript{1,}\thanks{\ \ Corresponding author.} \quad
        \bf Taesung Lee\textsuperscript{2}  \\
  \textsuperscript{*}Equal contribution \\
  \textsuperscript{1}University of Massachusetts Amherst \quad
  \textsuperscript{2}Anthropic \\
  \texttt{\{shambhavikri,atharvashrik,cragarwal,sgovindan\}@umass.edu} \\
  \texttt{hschang@cs.umass.edu, elca4u@gmail.com}
}

\begin{document}
\maketitle
\begin{abstract}
Large language models are increasingly deployed across diverse applications. This often includes tasks LLMs have not encountered during training.
This implies that enumerating and obtaining the high-quality training data for all tasks is infeasible. Thus, we often need to rely on transfer learning using datasets with different characteristics, and anticipate out-of-distribution requests.
Motivated by this practical need, we propose an analysis framework, building a transfer learning matrix and dimensionality reduction, to dissect these cross-task interactions.
We train and analyze 10 models to identify latent abilities (e.g., Reasoning, Sentiment Classification, NLU, Arithmetic)
and discover the side effects of the transfer learning.
Our findings reveal that performance improvements often defy explanations based on surface-level dataset similarity or source data quality. Instead, hidden statistical factors of the source dataset, such as class distribution and generation length proclivities, alongside specific linguistic features, are actually more influential.
This work offers insights into the complex dynamics of transfer learning, paving the way for more predictable and effective LLM adaptation.
\end{abstract}

\section{Introduction}

Large Language Models (LLMs) demonstrate remarkable capabilities across diverse tasks, yet their deployment in real-world applications faces significant practical constraints. Cost and latency considerations render giant all-purpose models impractical for many use cases, driving widespread adoption of task-specific fine-tuning. However, this approach encounters a fundamental challenge: high-quality training data for target tasks is often unavailable or proprietary. Moreover, deployed LLMs routinely face out-of-distribution (OOD) requests that extend beyond their fine-tuning scope.  This is especially true for agentic systems, which rely heavily on cross-domain skill transfer to perform diverse sequences of tasks. These realities necessitate a deeper understanding of transfer learning.

\begin{figure}[t!]
    \centering
    \includegraphics[width=1\columnwidth]{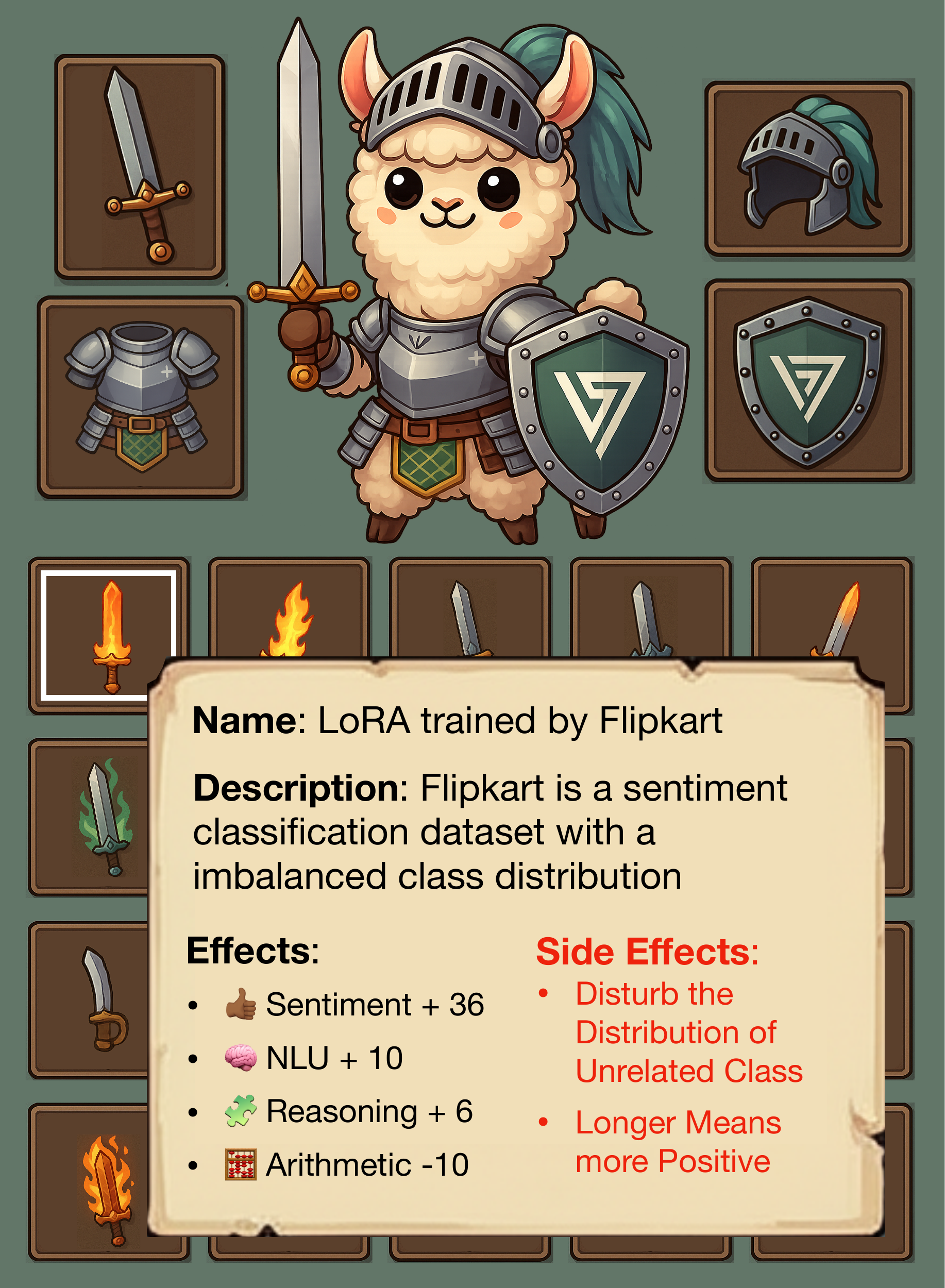} 
    \caption{Illustration of our motivations. LLMs such as Llama can be equipped with many different performance enhancers such as LoRA fine-tuned on a specific dataset. Our goal is to discover the potential impacts on out-of-domain tasks and side effects of each equipment.}
    \label{fig:first_fig}
\end{figure}

Traditional transfer learning research has primarily focused on scenarios where source and target tasks share the same domain, assuming that similar or related in-domain data provides useful signal for the target task.
However, the diverse task landscape that modern LLMs navigate demands deeper understanding of OOD transfer effects. Our experiments using Low-Rank Adaptation (LoRA) reveal counterintuitive transfer behaviors: fine-tuning on one dataset can yield surprising performance improvements or degradations on seemingly unrelated tasks, often defying expectations based on semantic similarity (illustrated conceptually in \Cref{fig:first_fig}). This unpredictability creates significant challenges for practitioners selecting optimal source datasets for fine-tuning, particularly in resource-constrained environments where training efficiency is paramount, or when acquiring pre-trained LoRA adapters from service providers without clear transferability guarantees.

In this paper, we propose a framework to analyze how the source fine-tuning dataset influences the performances on the target datasets in transfer learning and use this framework to systematically characterize the OOD generalization of an LLM using multiple LoRA adapters. Our analysis framework first constructs a performance matrix across different source and target tasks. We apply Principal Component Analysis (PCA) to this matrix to uncover latent abilities or "traits" that fine-tuned LLMs acquire from the transfer learning. We demonstrate that straightforward factors like source data quality or simple source-target similarity often fail to explain observed transfer learning effects.
Instead, we highlight the critical role of more subtle, "hidden" statistical properties of the source training data (e.g., output length distribution, label imbalance) and learned sensitivities to specific linguistic features.

Our work aims to provide actionable insights into the selection of the source dataset for fine-tuning, fostering a deeper understanding of the interactions among the datasets and guiding the development of more robust LLM adaptation strategies. 
In our experiments, we fine-tune the Llama 3.2 3B base model \cite{DBLP:journals/corr/abs-2407-21783} using LoRA and systematically evaluate models fine-tuned on one dataset across datasets for math, coding, natural language inference, sentiment, and toxicity detection tasks to map diverse data interactions. Through analyzing the fine-tuned LLM and datasets, we observe several surprising cross-domain interactions, including: (1) the impact of source data generation length on fine-tuned model outputs;
(2) asymmetric enhancement through out-of-domain fine-tuning datasets; and (3) the profound effects of source label imbalance on both in-domain and OOD performance.

\section{Related Work}

The transfer learning of fine-tuning language models is investigated by several existing works
\citep{vu2020exploringpredictingtransferabilitynlp,chang2021rethinking,parvez2021evaluating,weller2022usemultitasklearningvs,padmakumar2022exploring,li2024identifying,schulte2024less,yang2024unveiling,li2024scalable}. Most studies focus on identifying similar tasks for positive transfer effect through fully fine-tuning small language models. Instead, our work focuses on modeling the impact of LoRA fine-tuning and discovering the often-overlooked side effect of the source training datasets including out-of-domain and out-of-distribution datasets. Compared to the full fine-tuning, \citet{lin2024lora, zhou2024closer} find LoRA ``learns less and forgets less'', which potentially preserves out-of-domain base model capabilities better. This is one of the main reasons behind LoRA's effectiveness and popularity. Nevertheless, we demonstrate that LoRAs, which are fine-tuned on many source datasets, could still cause several types of undesirable side effects when being evaluated on a wide range of target tasks.

Methodologically, our analysis framework is related to \citet{sorensen2024observational}, which employs PCA to analyze observational scaling laws and the predictability of LLM performance across different model sizes and tasks. Some recent findings also support our discoveries of hidden factors. For example, \citet{yuan2024unveilingcoding} report that instruction fine-tuning with coding data can sometimes negatively impact mathematical reasoning. \citet{guha2025openthoughtsdatarecipesreasoning} find that the length distribution of the instruction tuning training data could affect the LLMs' code generation ability. \citet{min-etal-2022-rethinking,kung2023models,guha2025openthoughtsdatarecipesreasoning} discover that the format of the fine-tuning data might be more important than its content or correctness. Our work confirms their findings and provides a more comprehensive list of latent traits that influence LoRAs' performance.

\begin{figure*}[t!]
    \centering
    \includegraphics[width=\textwidth]{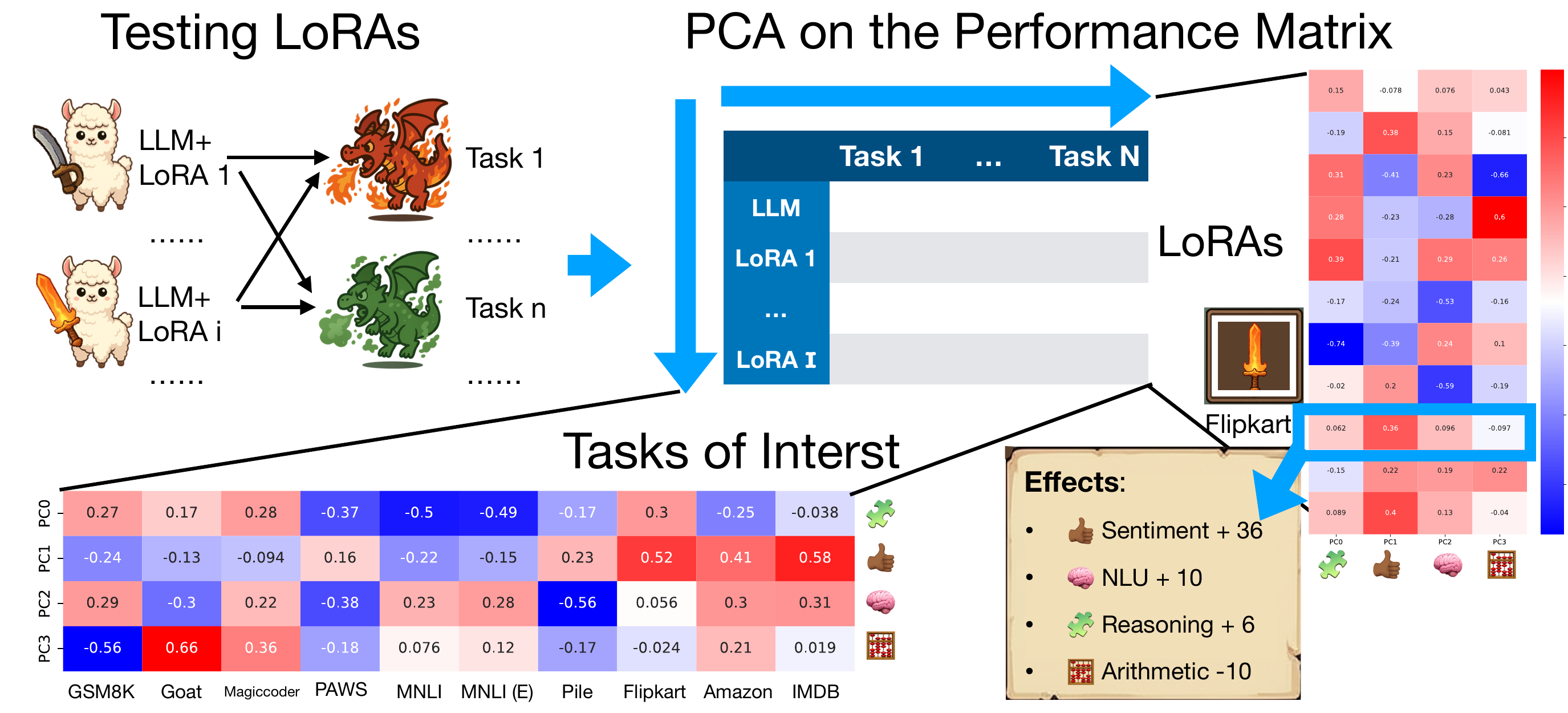} 
    \caption{Discovering the latent traits of LoRA through PCA. The performance matrix stores the performance of $I$ LoRAs on $N$ tasks. A PCA factorizes the performance matrix into two matrices: the top four eigenvectors/bases in the bottom left and the linear weights that combine the eigenvectors/bases in the right. More red means the values are higher. Based on the eigenvectors, we identify the meaning of each principal component as our latent traits, and we can use the linear weights of the LoRA trained on Flipkart as its influence to the other datasets through the traits.}
    \label{fig:pca_fig}
\end{figure*}

\section{Methodology}

In our framework, we first prepare $N$ representative tasks/datasets of interest and fine-tune LLMs on these $N$ datasets to acquire $I$ fine-tuned LLM variants. In this paper, we use LoRA to fine-tune LLMs, but the framework could be applied to any fine-tuning method (e.g., full fine-tuning, prompt tuning~\citep{lester2021power}, BitFit~\citep{zaken2022bitfit}, etc.) or any variants of LLMs (e.g., in-context learning or chain of thoughts).

Viewing each fine-tuned LoRA adapter as a specialized piece of equipment in a practitioner's toolkit, a crucial challenge is selecting the right tool for a new task. The conventional approach assumes a tool's effectiveness is dictated by its labeled domain, for instance, using a `sentiment' adapter for a sentiment task. However, these tools might come with unexpected side-effects and hidden capabilities driven by the latent statistical properties of their training data, not just their domain.

To solve this issue, we evaluate the $I$ fine-tuned LLMs on the $N$ datasets on their accuracy, and organize the pairwise results into a $I \times N$ performance matrix as shown in \Cref{fig:pca_fig}.
Note that since our goal is to measure the impact on out-of-domain tasks, we focus more on relative gains, rather than the absolute performance numbers.

Throughout this paper, we denote the base LLM as \textit{M}; a model fine-tuned on a dataset \textit{D} as \textit{M}["D"]; and the output performance of such a model on the evaluation data X is denoted \textit{M}["D"](X). For example, \textit{M}["Flipkart"](GSM8K) refers to the score of the model fine-tuned on the Flipkart dataset and tested on GSM8K. In the performance matrix, \textit{M}["Flipkart"](GSM8K) corresponds to the row for Flipkart and column for GSM8K.

To understand the overall characteristics and transfer learning impact across these datasets, we decompose the performance matrix using PCA. Each principal component corresponds to a group and the tasks with high values in the corresponding eigenvector belong to the group. In this way, similar evaluation tasks whose LLM scores have high correlations will cluster together. We can then use the common attribute of the tasks in a group as its name - a standard practice to make the abstract mathematical components interpretable. Guided by the PCA results, we discover the transfer learning patterns among the tasks of interest and further investigate the outliers in the performance matrix. We then conduct analyses to identify the factors that could explain the patterns and outliers.

\begin{table*}[t!]
\centering
\small
\resizebox{\textwidth}{!}{%
\begin{tabular}{lcccccccccc}
\toprule
\textbf{Fine-tuned on} & \textbf{GSM8K} & \textbf{Goat} & \textbf{Magicoder} & \textbf{PAWS} & \textbf{MNLI} & \textbf{MNLI (E)} & \textbf{Pile} & \textbf{Flipkart} & \textbf{Amazon} & \textbf{IMDB} \\
\midrule
None (Original LLM) & 9.78  & 6.36  & 21.55  & 46.30  & 33.30  & 33.75  & 38.45  & 63.55  & 31.80  & 51.45  \\
MetaMath          & 44.96 & 5.40  & 20.50  & 44.55  & 34.65  & 32.95  & 42.25  & 46.10  & 21.60  & 49.65  \\
Goat              & 13.42 & 24.65 & 21.65  & 44.55  & 33.10  & 35.10  & 47.15  & 57.85  & 25.45  & 53.00  \\
Magicoder         & 19.18 & 8.45  & 29.38  & 45.70  & 33.35  & 34.10  & 37.35  & 70.90  & 28.10  & 51.15  \\
PAWS              & 8.57  & 7.75  & 20.75  & 70.05  & 34.90  & 33.15  & 49.05  & 12.10  & 21.85  & 46.80  \\
MNLI (Eng.)       & 9.33  & 6.00  & 20.55  & 57.65  & 69.50  & 83.45  & 51.10  & 5.85   & 37.70  & 53.50  \\
Pile              & 12.66 & 8.16  & 21.47  & 56.35  & 35.70  & 33.65  & 85.25  & 83.90  & 32.80  & 51.00  \\
Flipkart          & 14.59 & 5.96  & 21.84  & 55.55  & 33.65  & 36.25  & 49.10  & 92.65  & 38.70  & 77.15  \\
Amazon            & 12.97 & 9.15  & 22.48  & 55.45  & 39.10  & 38.35  & 47.90  & 39.95  & 61.25  & 69.05  \\
IMDB              & 12.78 & 6.96  & 22.19  & 55.40  & 34.00  & 34.70  & 46.35  & 85.55  & 31.40  & 91.45  \\
\bottomrule
\end{tabular}%
}
\caption{Model fine-tuning and cross-task evaluation results (\% Automatic Accuracy or Accuracy from LLM-as-a-Judge). Each model was fine-tuned on a single dataset (leftmost column) and evaluated across multiple target tasks (column headers). MNLI (E) refers to MNLI English.}
\label{tab:full-train-test_results}
\end{table*}

\section{Experimental Setup}

We curate a diverse set of datasets spanning mathematical reasoning (MetaMath \cite{yu2023metamath}, GSM8K \cite{cobbe2021training}, and Goat \cite{liu2023goat}), code generation (Magicoder \cite{ise_uiuc_2023}), Natural Language Inference (NLI) (PAWS \cite{zhang2019pawsparaphraseadversariesword} and MNLI \cite{N18-1101}), Sentiment analysis (Flipkart Sentiment \cite{kayee_flipkart_sentiment_analysis}, Amazon Reviews \cite{xiang_zhang_acharki_yassir_2022}, and IMDB Reviews \cite{maas-EtAl:2011:ACL-HLT2011}), and toxicity detection (Pile \cite{korbak2024piletoxicity}). For more information about the datasets refer to \Cref{tab:datasets_overview} in Appendix~\ref{app:dataset}.

We employ Low-Rank Adaptation (LoRA) with rank 64 to fine-tune the Llama 3.2 3B base model \textit{M} on each source dataset to get a fine-tuned \textit{M}[Dataset]. \footnote{MetaMath is designed for training, so we replace MetaMath with GSM8K in evaluation.}
For all tasks, we report the accuracy using LLM-as-a-Judge. \footnote{While widely used for scalable evaluation, we acknowledge that the LLM-as-a-Judge method may introduce its own inherent biases, a potential limitation of our evaluation framework.}  Specifically, we use Llama 3.3 70B Instruct \cite{DBLP:journals/corr/abs-2407-21783} to judge if the generated answers are the same as the ground truth answer (see Appendix \ref{sec:appendix} for prompt). 
For each dataset, 10,000 samples are randomly chosen from its training split, and 2,000 from the test split unless specified otherwise. Model training specifics are detailed in Section~\ref{app:implementation}.

\section{Results and Analyses}
In this section, we show the analysis on how the statistical properties drive transfer learning regardless of the domain similarity.
We show various statistical properties and their effects on performance for both in-domain transfer and out-of-domain transfer.
The overall cross-task performance matrix is summarized in \Cref{tab:full-train-test_results}.

\subsection{PCA Results}

The results of PCA on the performance matrix are visualized in \Cref{fig:pca_fig}. The first four eigenvectors, which explain around 75\% of the total variance in the performance matrix, are presented at the bottom-left of the figure and each column corresponds to a target evaluation task in \Cref{tab:full-train-test_results}. 

The first principal component (PC0) assigns positive values to GSM8K, Goat, Magicoder, and Flipkart, suggesting that PC0 measures the \textit{reasoning} performance of LoRAs. Surprisingly, Flipkart is also included in the group. The second principal component (PC1) group consists of PAWS, Pile, Flipkart, Amazon, and IMDB, which are mostly \textit{sentiment classification} datasets except for PAWS. The PC2 groups GSM8K, Magicoder, MNLI, MNLI (E), Amazon, and IMDB together, so we believe the group represents the general \textit{natural language understanding} (NLU) performance. 
Finally, PC3 highlights the performance differences between GSM8K and Goat. \Cref{tab:full-train-test_results} shows that LoRA fine-tuned on MetaMath actually decreases the performance on Goat. We hypothesize that this is because Goat tests the arithmetic for large numbers while GSM8K only requires the arithmetic for small numbers. Thus, we annotate PC3 as LoRAs' ability of performing \textit{arithmetic} for large numbers due to its large positive value to Goat. The positive values of Magicoder and Amazon might indicate that solving these tasks also require this arithmetic skill.

According to our annotation of every principal component, we can characterize LoRA fine-tuned by every source dataset based on the values projected to each principal component. For example, the table on the right side of \Cref{fig:pca_fig} shows that LoRA from the Flipkart sentiment classification task improves sentiment ability the most as expected. Besides, it also slightly improves the NLU and reasoning ability of LLMs while degrading the arithmetic performance.

\begin{figure}[t!]
    \centering
    \includegraphics[width=1.0\columnwidth]{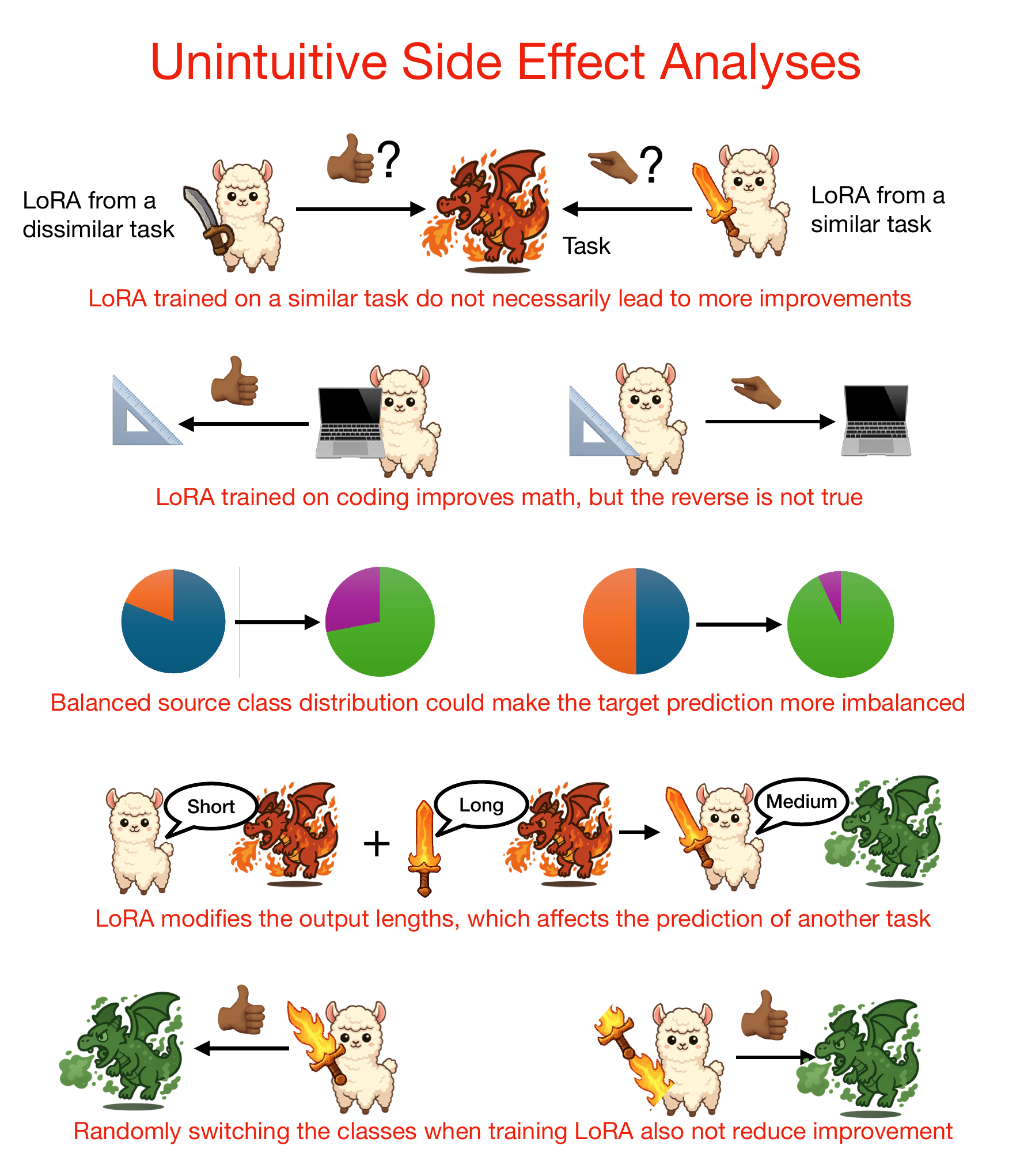} 
    \caption{Unintuitive side effects of using LoRA adapters as specialized `tools'. This figure illustrates surprising behaviors where a tool's performance is not predicted by its label: domain similarity can be misleading, skill transfer is often asymmetric, and hidden statistical properties like class balance and output length proclivities are transferred to new tasks with unexpected consequences.}
    \label{fig:side_effect_fig}
\end{figure}

\subsection{Analyzing Side Effects of Cross-Task Transfer Systematically}

To map the behaviors of transfer learning, we categorize them using the $2 \times 2$ table in Table~\ref{tab:transfer_framework}. This table helps explain the counterintuitive results observed in our experiments: why a LoRA trained on the tasks from a different domain (e.g., a `code generation' adapter) might surprisingly outperform an in-domain one for a specific mathematical task, or why two seemingly identical `sentiment' LoRAs can have vastly different effects on the target task.
The following sections will deconstruct the specific properties of these LoRAs, analyzing their generation length proclivities (\Cref{subsec:len_dist}), internal class distributions (\Cref{subsec:class_dist}), learned linguistic sensitivities (\Cref{subsec:ling_transfer}), and the correctness of the labels (\Cref{sec:mislabel}) to explain the surprising dynamics. We illustrate the most notable side effects in \Cref{fig:side_effect_fig}.

\begin{table}[!th]
\centering
\small
\resizebox{\columnwidth}{!}{%
\begin{tabular}{l p{3.5cm} p{3.5cm}}
\toprule
& \textbf{Same Domain} & \textbf{Different Domain} \\
\midrule
\textbf{Different Stats} &
\textbf{Unexpected Negative Transfer}
\begin{itemize}[nosep, leftmargin=*]
    \item Amazon \textrightarrow{} Flipkart (both sentiment) shows poor transfer.
    \item Flipkart (balanced vs. imbalanced) yields divergent results on other sentiment tasks.
\end{itemize} &
\textbf{Asymmetric \& Negative Transfer}
\begin{itemize}[nosep, leftmargin=*]
    \item Asymmetric Transfer: Code \textrightarrow{} Math (+9.4) but Math \textrightarrow{} Code (-1.05).
    \item Math \textrightarrow{} Sentiment shows strong negative transfer (e.g., Flipkart, -17.45).
\end{itemize} \\
\midrule
\textbf{Similar Stats} &
\textbf{Traditional Expectation}
\begin{itemize}[nosep, leftmargin=*]
    \item IMDB \textrightarrow{} IMDB shows strong in-domain performance (91.45\%).
    \item MNLI \textrightarrow{} MNLI (E) is also strong (83.45\%).
\end{itemize} &
\textbf{Surprising Positive Transfer}
\begin{itemize}[nosep, leftmargin=*]
    \item \textit{Length Similarity:} Code \textrightarrow{} Math transfer outperforms in-domain Math \textrightarrow{} Math.
    \item \textit{Linguistic Transfer:} Classification \textrightarrow{} Math improves reasoning.
\end{itemize} \\
\bottomrule
\end{tabular}%
}
\caption{A summary for cross-task side effects.}
\label{tab:transfer_framework}
\end{table}

\subsection{Length Distribution}
\label{subsec:len_dist}
\begin{figure}[t!]
    \centering
    \includegraphics[width=\columnwidth]{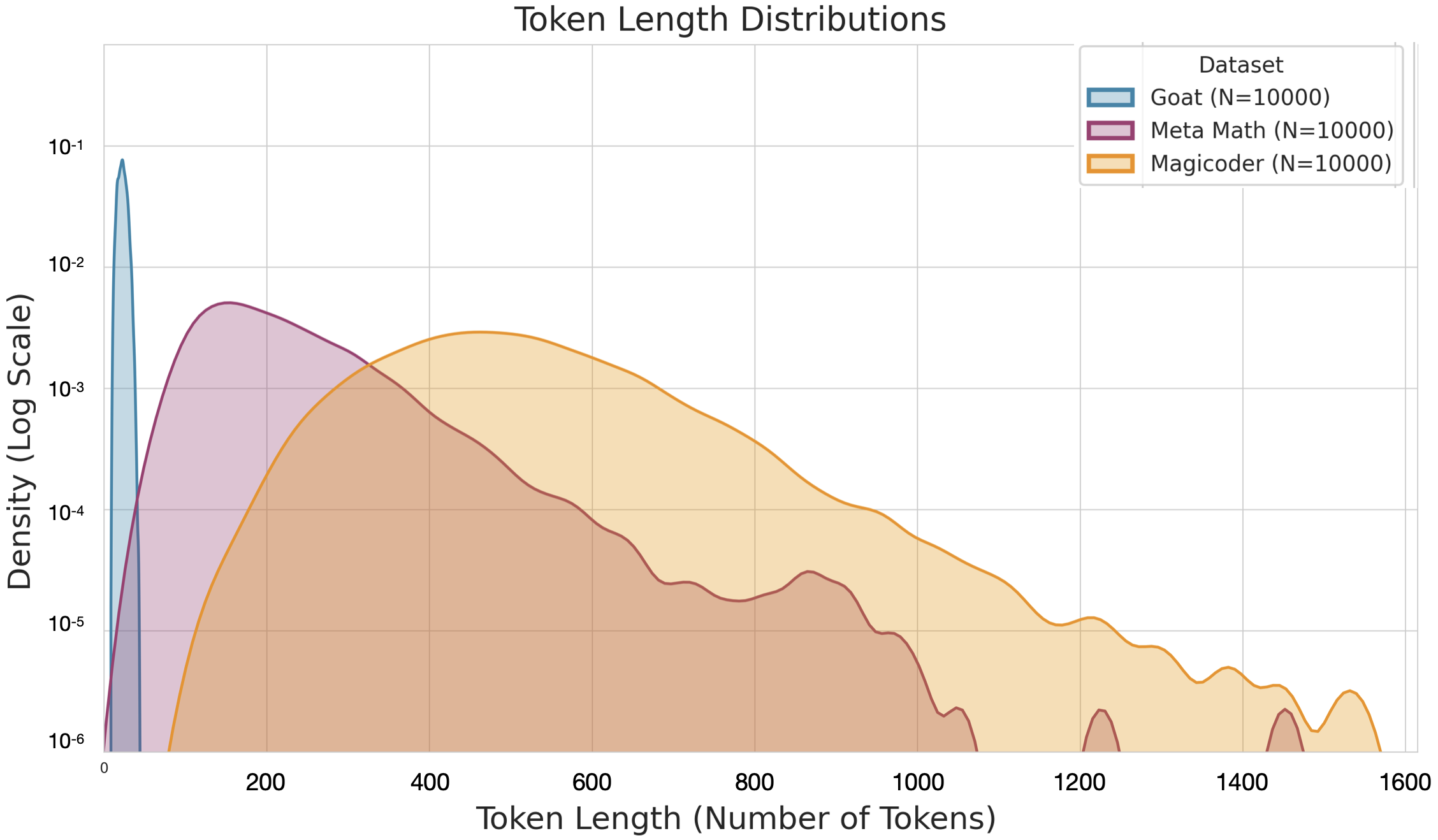}
    \caption{Generation length differences across Meta Math, Goat and Magicoder datasets.}
    \label{fig:gsm-vs-goat}
\end{figure}

We observe that performance changes sometimes align with the length distributions of the fine-tuning and evaluation datasets, a characteristic learned by the model that influences output length on the target task.
For example, while both Meta-Math/GSM8K and Goat are Math domain datasets, Goat has a significantly shorter generation length distribution (Figure~\ref{fig:gsm-vs-goat}).
Fine-tuning on Magicoder, a code dataset with a length distribution more similar to Meta-Math/GSM8K's, proved more effective on GSM8K (+9.40 gain) than fine-tuning on the in-domain Goat dataset (+3.64 gain). This suggests that matching generation length proclivities can be crucial for positive transfer.

However, this phenomenon is sophisticated and influenced by several interacting factors (detailed in Appendix \ref{app:len_dist}):
\begin{itemize}
    \item \textbf{Interpolation of Lengths:} Models fine-tuned on generation tasks often produce outputs whose lengths interpolate between the base model's tendencies and those of the fine-tuning data.
    \item \textbf{Classification Task Influence:} Fine-tuning on classification datasets generally preserves the base model's generation length on OOD generation tasks, unless the classification data itself has a strong length bias.
    \item \textbf{Dataset-Specific Length Transfer:} Certain datasets (e.g., Pile) can impart distinct length tendencies that transfer to OOD tasks.
    \item \textbf{Length Bias in Classification Inputs:} Correlations between input text length and class labels in a source classification dataset can be learned and transferred, affecting predictions on target classification tasks. 
\end{itemize}

These findings suggest that generation length is a transferable latent trait. Models exhibit a form of ``inertia'', blending prior generation habits with newly learned ones from the fine-tuning data. This has implications for multi-task learning, as unintended output lengths could affect downstream performance or introduce subtle biases.

\subsection{Class Distribution}
\label{subsec:class_dist}
In classification tasks, the model needs to learn the features of the input and predict a series of tokens representing a class. We find that fine-tuning can shift this output class distribution in unexpected ways for both in-domain and out-of-domain tasks. Notice that when analyzing the class distributions, we can often ignore the impact of the length distribution because the outputs of the classification tasks are typically only a couple of tokens.

\begin{figure}[!t]
    \centering
    \includegraphics[width=\columnwidth]{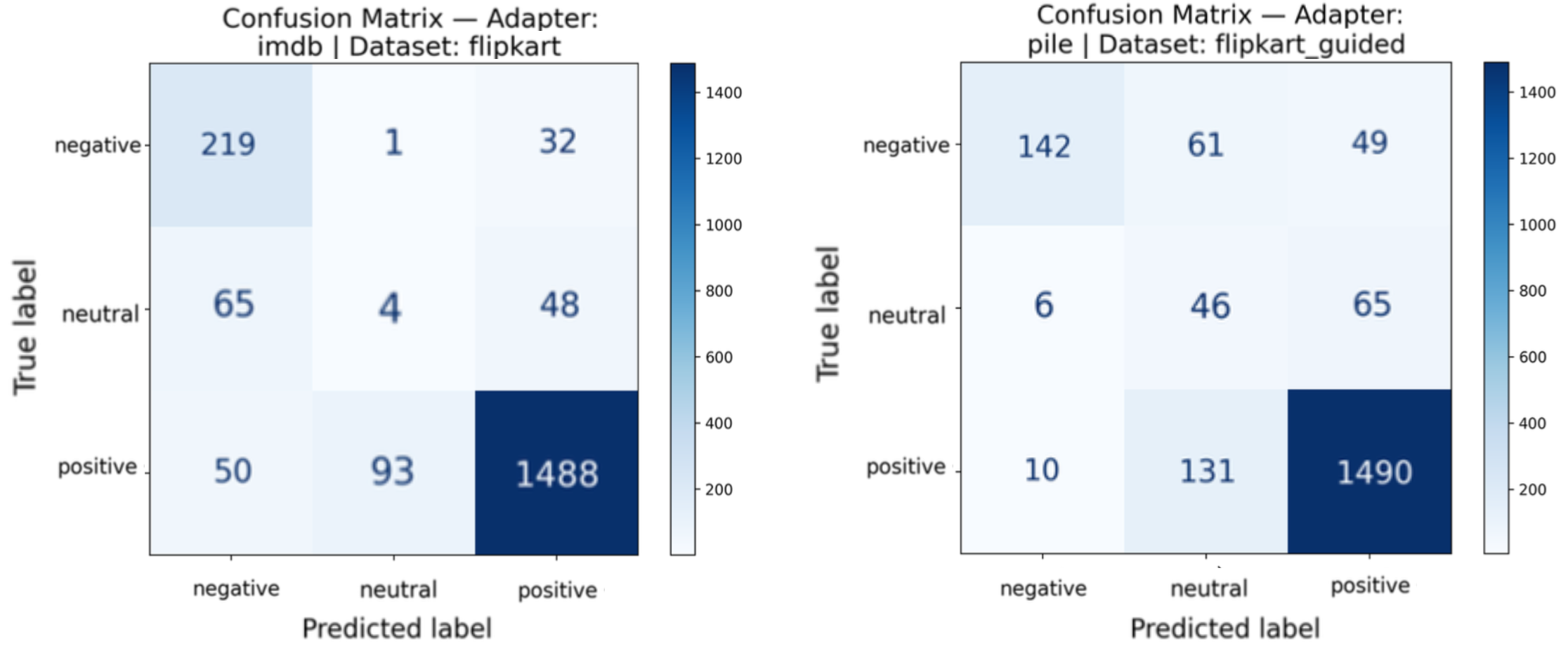} 
    \caption{Confusion Matrices on Flipkart: \textit{M}[IMDB] (left) vs. \textit{M}[Pile] (right).}
    \label{fig:cm2_results}
\end{figure}

\begin{figure}[t!]
    \centering
    \includegraphics[width=\columnwidth]{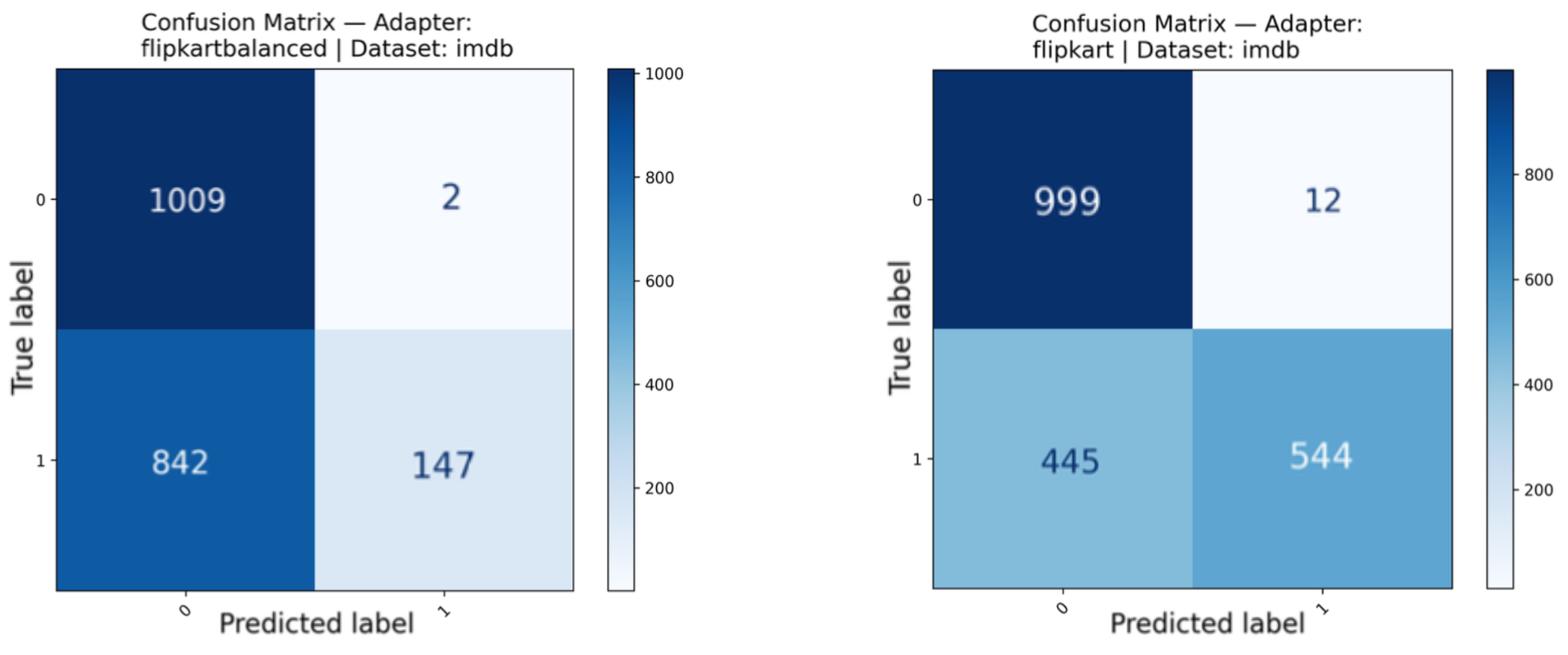} 
    \caption{Confusion Matrices on IMDB (binary): \textit{M}[Flipkart(Balanced)] (left) vs. \textit{M}[Flipkart(Imbalanced)] (right) : 0=negative, 1=positive.}
    \label{fig:cm_flipkart_imdb_combined_results}
\end{figure}

\begin{figure*}[!th]
    \centering
    \includegraphics[width=1.0\textwidth]{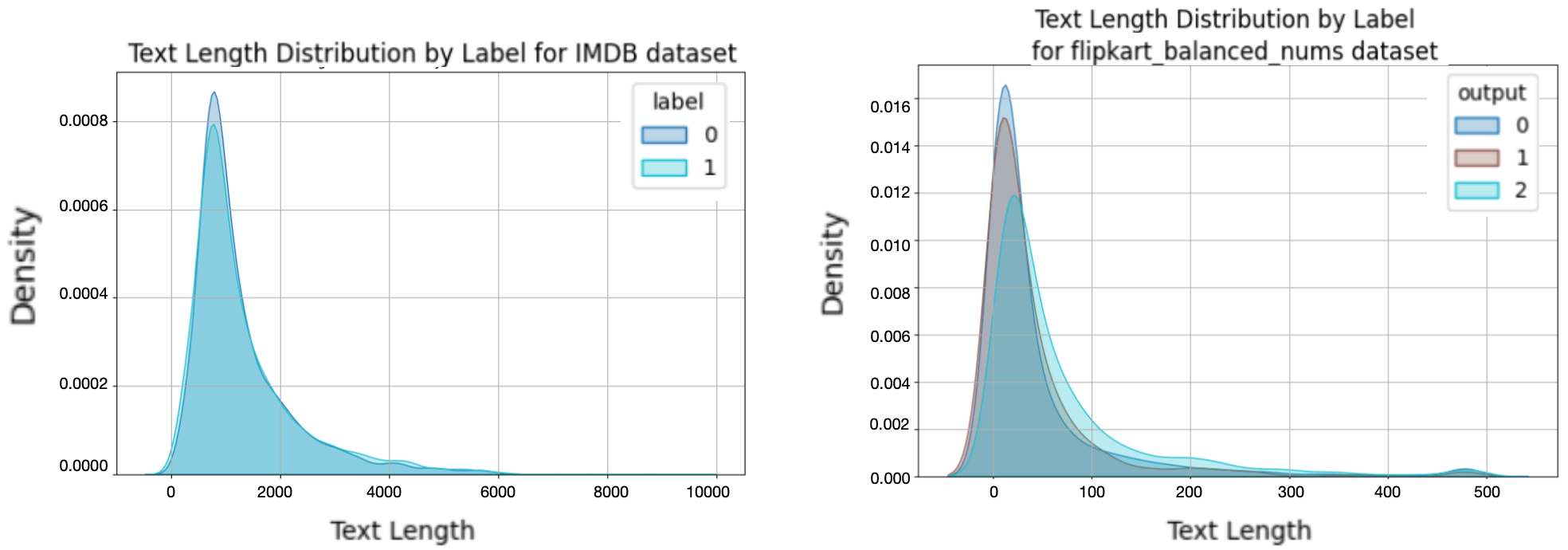} 
    \caption{Text length distribution of each sentiment label from the kernel density estimation (KDE) for IMDB (left: 0=negative, 1=positive) and Flipkart (right: 0=positive, 1=neutral, 2=negative). }
    \label{fig:lens_8b_results}
\end{figure*}

With the high similarity between the classification tasks, we could observe positive transfer between classification tasks for many dataset pairs (Table \ref{tab:label_imbalance_subset_results}). For example, \textit{M}[Pile](Flipkart) improves performance to 83.90\%, and \textit{M}[IMDB](Flipkart) improves to 85.55\%, as compared to 63.55\% from \textit{M}(Flipkart). Moreover, we observed that prediction bias could be learned and applied to a different task, both in domain and across domains.
For instance, \Cref{fig:cm2_results} shows that \textit{M}[Pile] predicts `neutral' more often on Flipkart than \textit{M}[IMDB], which suggests that training on Pile (toxicity) might increase sensitivity to ambiguous language, while IMDB training (binary sentiment) pushes for definitive positive/negative calls.

To further isolate the effect of label distribution from the task itself, we increase the negative class ratio from around 20\% to 50\%. The newly created dataset is called Flipkart-balanced, while the original Flipkart is called Flipkart-imb. Comparing LoRA \textit{M}[Flipkart-balanced] with \textit{M}[Flipkart], \Cref{tab:label_imbalance_subset_results} highlights target-dependent effects due to the class distribution similarity and the dissimilarity between the fine-tuning and evaluation datasets. \textit{M}[Flipkart-balanced](Pile) performs better than \textit{M}[Flipkart-imb.](Pile)(50.05\% vs. 39.10\%), while \textit{M}[Flipkart-imb.](IMDB) is better (77.15\% vs. 57.80\%). Balancing may help tasks needing unbiased signals (toxicity - Pile), while natural imbalance can preserve useful priors for OOD tasks with similar distributions (sentiment - IMDB).

\Cref{fig:cm_flipkart_imdb_combined_results} compares \textit{M}[Flipkart-imb.](IMDB) and \textit{M}[Flipkart-balanced](IMDB), which demonstrates a bias towards predicting `negative', especially \textit{M}[Flipkart-balanced]. This might be linked to learning spurious features like the input length and predicting long inputs as negative because negative reviews in Flipkart are longer than positive reviews, unlike IMDB's more uniform lengths as shown in \Cref{fig:lens_8b_results}.

\begin{table}[t!]
\centering
\small
\resizebox{\columnwidth}{!}{%
\begin{tabular}{lrrrr}
\toprule
\textbf{Model (FT on)} & \textbf{Pile} & \textbf{Flipkart} & \textbf{Amazon} & \textbf{IMDB} \\
\midrule
Original LLM       & 38.45 & 63.55 & 31.80 & 51.45 \\
Pile           & 85.25 & 83.90 & 32.80 & 51.00 \\
Flipkart-imb. & 39.10 & 92.65 & 38.70 & 77.15 \\
Flipkart-bal. & 50.05 &  N/A  & 38.80 & 57.80 \\
Amazon         & 47.90 & 39.95 & 61.25 & 69.05 \\
IMDB           & 46.35 & 85.55 & 31.40 & 91.45 \\
\bottomrule
\end{tabular}%
}
\caption{Label Imbalance Effects on Classification Tasks. Flipkart-bal. means Flipkart-balanced, Flipkart-imb, means Flipkart-imbalanced.}
\label{tab:label_imbalance_subset_results}
\end{table}

\subsection{Transferring from Classification to Math}
\label{subsec:ling_transfer}

Fine-tuning on classification datasets shows a surprising ability to improve performance on mathematical reasoning tasks, particularly GSM8K. \Cref{tab:classification_ft_math_results} shows that several LoRAs trained on classification tasks improved GSM8K accuracy over the original LLM. For example, \textit{M}[Flipkart] achieves 14.59\%, which is much higher than 9.78 from \textit{M}. This gain was less pronounced on Goat, a more arithmetic-focused dataset, suggesting the improvement is more related to linguistic reasoning than raw calculation.

\begin{table}[t!]
\centering
\small
\resizebox{\columnwidth}{!}{%
\begin{tabular}{lcc}
\toprule
\textbf{Model Fine-tuned on} & \textbf{GSM8K Acc. (\%)} & \textbf{Goat Acc. (\%)} \\
\midrule
None (Original LLM)             & 9.78     & 6.36   \\
Flipkart (Imbalanced) & 14.59    & 5.96   \\
Flipkart (Balanced)  & 13.00    & 6.50   \\ 
Amazon               & 12.97    & 9.15   \\
IMDB                 & 12.78    & 6.96   \\
Pile                 & 12.66    & 8.16   \\
PAWS                 & 8.57     & 7.75   \\
MNLI (Eng.)          & 9.33     & 6.00   \\
\bottomrule
\end{tabular}%
}
\caption{Performance of models fine-tuned on classification datasets, evaluated on GSM8K and Goat.}
\label{tab:classification_ft_math_results}
\end{table}

\begin{figure}[!t]
    \centering
    \begin{subfigure}[b]{0.48\textwidth}
        \includegraphics[width=\columnwidth]{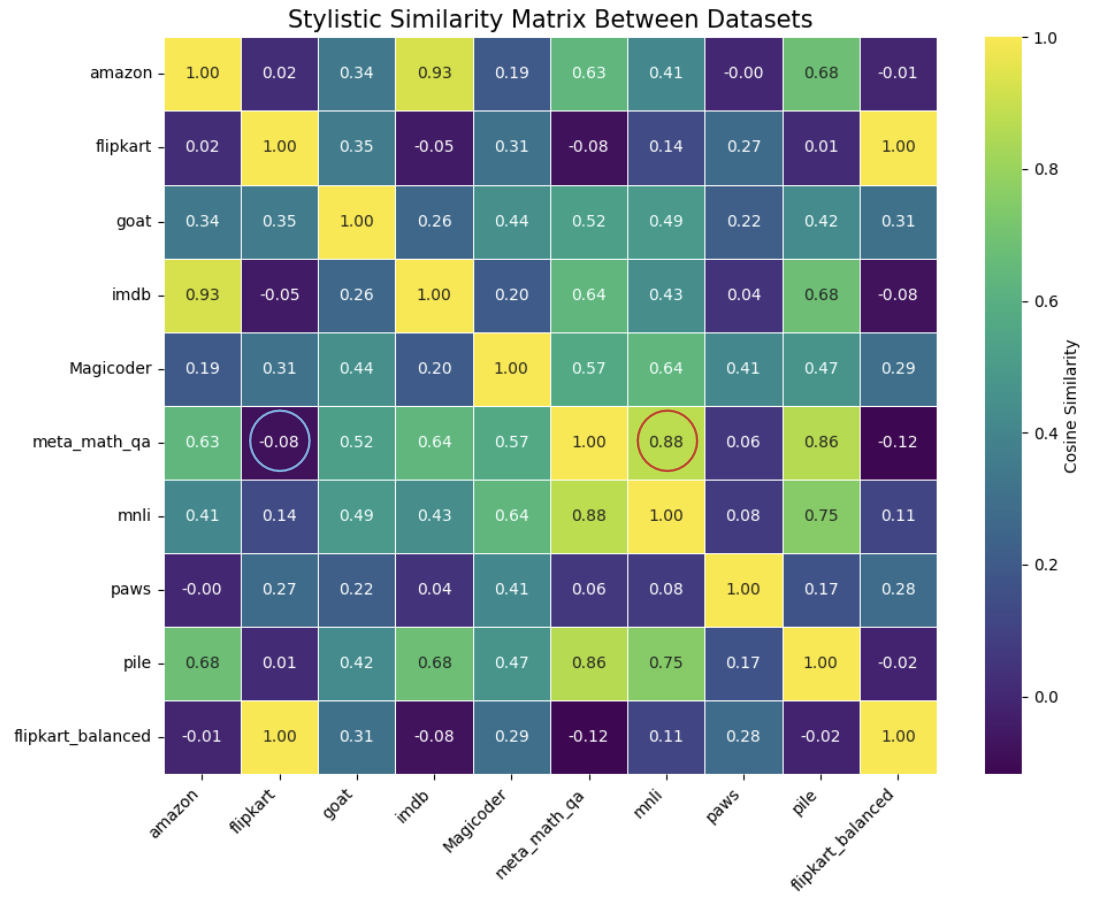} 
        \caption{Stylistic Similarity Matrix Between Datasets.}
        \label{fig:stylistic_similarity}
    \end{subfigure}
    \hfill
    \begin{subfigure}[b]{0.48\textwidth}
        \includegraphics[width=\columnwidth]{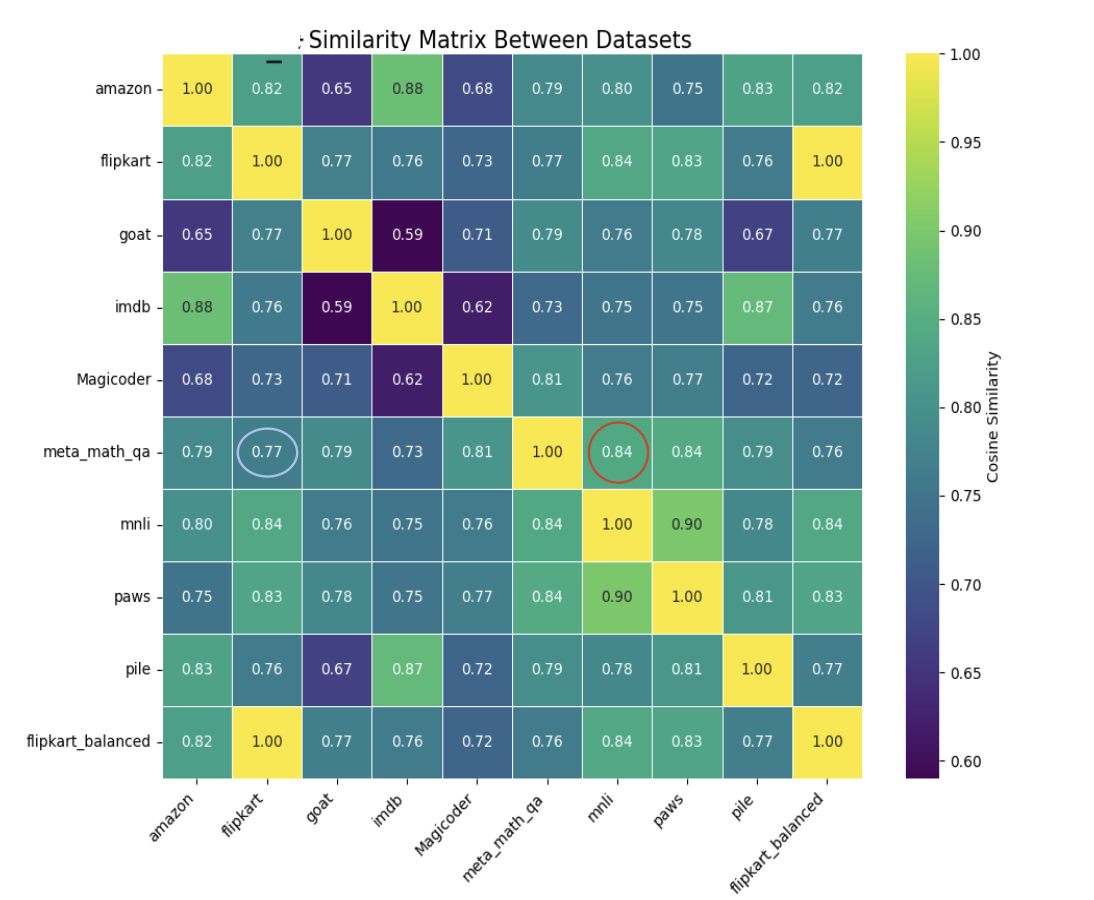}
        \caption{Semantic Similarity Matrix Between Datasets.}
        \label{fig:semantic_similarity}
    \end{subfigure}
    \caption{Stylistic Similarity (top) and Semantic Similarity (bottom) Matrices.}
    \label{fig:similarity_matrices}
\end{figure}

One initial hypothesis was that overall stylistic similarity~\citep{wegmann2022same} or semantic similarity\footnote{MiniLM-L6-v2 from \url{https://www.sbert.net/}} between source classification datasets and target math datasets might predict transfer. However, these broad similarities did not consistently correlate with the observed improvements on GSM8K, indicating that more nuanced factors are at play. For example, \Cref{fig:similarity_matrices} highlights that MNLI has very high similarities with MetaMath while there is almost no positive transfer between them. In contrast, Flipkart could significantly improve GSM8K, while being stylistically very dissimilar to MetaMath.

\begin{figure}[!t]
    \centering
    \includegraphics[width=\columnwidth]{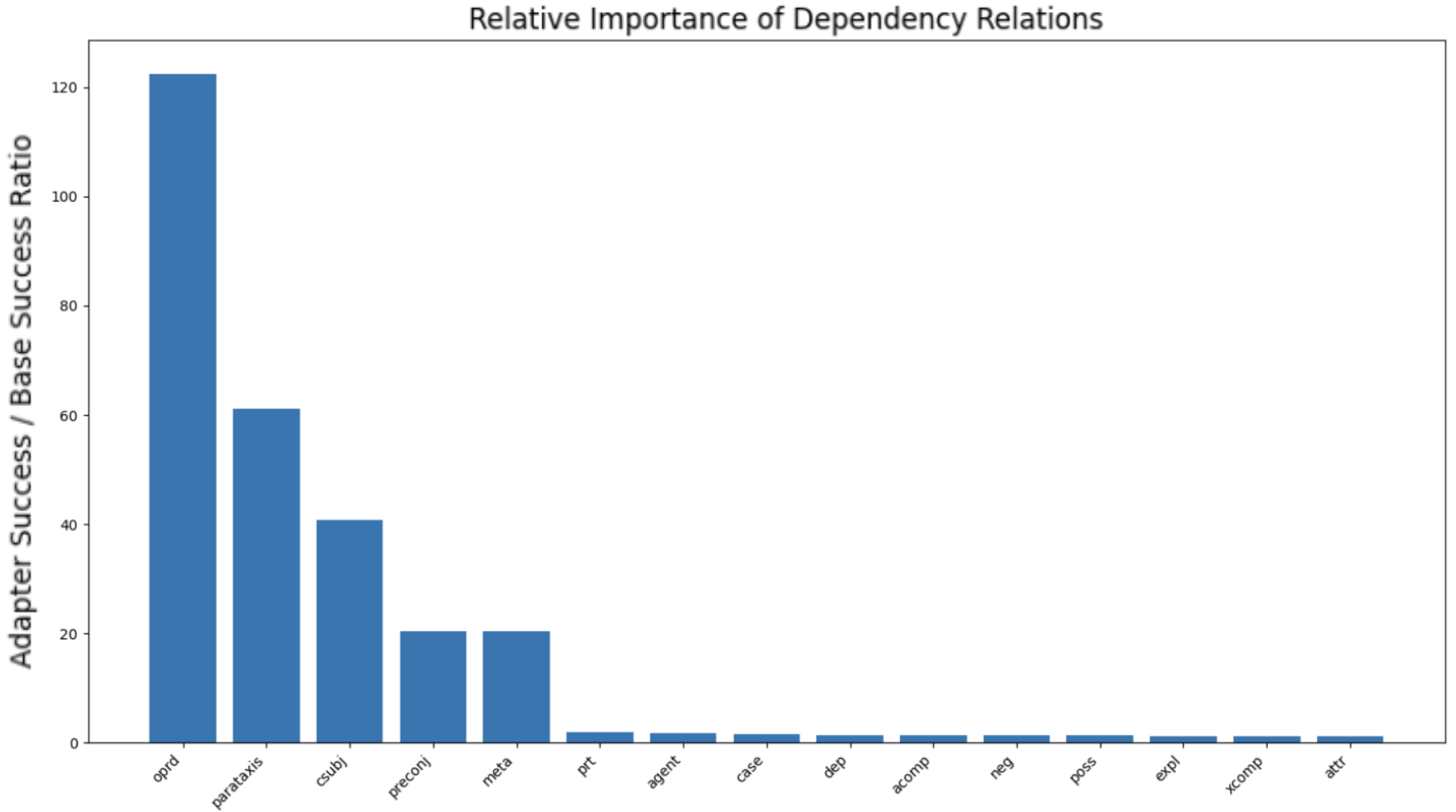} 
    \caption{Performance ratio (LoRA success / original LLM success) on math problems. Gains on relations like oprd show the model's improved ability to parse grammatical structure is the key driver of its success.}
    \label{fig:dependency_relations}
\end{figure}

Instead, our analysis (detailed in Appendix \ref{app:classification_math_details}) suggests that the improvement stems from an enhanced sensitivity to specific linguistic structures crucial for understanding and deconstructing word problems. Key observations include:
\begin{itemize}
    
    \item \textbf{Sensitivity to Syntactic Cues:} The fine-tuned LoRAs significantly improve the model's ability to interpret the grammatical structure of word problems, which is essential for translating text into correct mathematical operations. For example, \Cref{fig:dependency_relations} shows that the model becomes better at identifying dependency relations like oprd (operand), which flags a number as an object to be acted upon, and parataxis, which links related clauses together. This enhanced syntactic proficiency is not just a linguistic improvement; it is the mechanism that allows the model to more reliably deconstruct complex sentences into accurate logical and mathematical steps. A failure to parse these cues correctly often leads to building the wrong equation (e.g., adding numbers that should be multiplied).

    \item \textbf{Asymmetric Transfer:}
    Our analysis revealed a significant asymmetric skill transfer.
    For example, classification datasets such as Flipkart, Amazon and IMDB improve the performance on GSM8K (+4.81, +3.19, +3.00, respectively), but training on MetaMath did not improve on Flipkart, Amazon and IMDB (-17.45, -10.20, -1.80). Similarly,
    a strong positive transfer from code to math was observed, where \textit{M}[Magicoder] improved performance (+9.4) while the inverse resulted in -1.05.
    
    \item \textbf{Handling of Arithmetic Operations:} LoRAs fine-tuned on classification datasets lead to consistent gains in math reasoning, especially for high-frequency operations like addition, subtraction, and division (see Figure~\ref{fig:math_cluster} in the appendix). This improvement appears to be linked to the model's increased ability to attend to relevant tokens and avoid premature termination, leading to more complete reasoning chains.
\end{itemize}
These findings highlight that fine-tuning on classification tasks can, perhaps counter-intuitively, refine a model's linguistic processing in ways beneficial for structured reasoning tasks like mathematics, beyond what simple dataset similarity would predict.

\subsection{Importance of Labels}
\label{sec:mislabel}
To further understand what drives performance, especially the linguistic understanding gained during fine-tuning, we investigated the direct role of labels. We tested if LoRAs are sensitive to incorrect labels. Interestingly, fine-tuning on Amazon reviews where labels are deliberately mislabeled yielded similar or even slightly better performance on OOD tasks like IMDB and GSM8K (Table \ref{tab:mislabeled_amazon_results}), suggesting models can pick up underlying data structures even with noisy labels, or that the mislabeling process inadvertently created patterns beneficial for other tasks.

 \begin{table}[t!]
\centering
\small
\begin{tabular}{@{}lccc@{}}
\toprule
\textbf{Model (Fine-tuned on)} & \textbf{Flipkart} & \textbf{IMDB} & \textbf{GSM8K} \\
\midrule
\textit{M}[Amazon]        & 15.85\% & 48.65\% & 3.34\% \\
\textit{M}[Amazon-mislabeled] & 16.75\% & 48.80\% & 4.02\% \\
\bottomrule
\end{tabular}
\caption{Impact of Fine-tuning on Wrong Labels Compared to Correct Labels (\% Accuracy).}
\label{tab:mislabeled_amazon_results}
\end{table}

\section{Conclusion}
Our investigation into the cross-task dynamics of Low-Rank Adaptation (LoRA) confirms that transfer learning behavior is often unintuitive and defies explanations based on task domains or surface level dataset similarity. This work introduced a systematic framework to dissect these interactions, revealing that performance on target tasks is more influenced by the transfer of latent statistical and linguistic traits learned from a source dataset. We established the existence of these phenomena, such as asymmetric skill transfer and the impact of the class distribution, providing a new lens through which to view the fine-tuning process.

Our work paves the way for a more predictable and engineering-driven discipline around LLM adaptation. The logical next step is to move toward a modular approach, creating a "tool-belt" of skill adapters for agentic systems. An agent could then dynamically load a "conciseness adapter" for summarization or a "syntactic-parser adapter," like the one we observed emerging from classification data, for complex instruction understanding. While we demonstrated these dynamics on the Llama 3 architecture with LoRA, a critical next step is to validate and expand these findings across a broader range of models and adaptation methods to assess their scalability. By pursuing these avenues of modularity and prediction, we can build more robust and capable AI agents.


\section*{Acknowledgments}
We are also deeply grateful to the valuable feedback from Mengxue Zhang, Wenlong Zhao, Dhruv Agarwal, and Prof. Andrew McCallum. This work was supported in part by Center for Data Science and in part by the Center for Intelligent Information Retrieval. Any opinions, findings and conclusions or recommendations expressed in this material are those of the authors and do not necessarily reflect those of the sponsor.
\bibliography{custom}

\clearpage

\appendix
\section{Experimental Specifics}
\subsection{Dataset}
\label{app:dataset}
Table \ref{tab:datasets_overview} lists the datasets we considered for our analyses, ranging from generation to classification tasks, across different domains, class and length distributions, etc.

\begin{table*}[ht!]
\centering
\small
\begin{tabular}{@{}p{2.5cm}p{2cm}p{4.5cm}p{5.5cm}@{}}
\toprule
\textbf{Dataset} & \textbf{Domain} & \textbf{Characteristics} & \textbf{Class Labels / Distribution} \\ \midrule
MetaMath \cite{yu2023metamath} & Math & Multi-step reasoning & Generation \\
Goat \cite{liu2023goat} & Math & Short arithmetic & Generation (95\% inputs < 500 chars) \\
GSM8K \cite{cobbe2021training} & Math & Grade school math problems & Generation (75\% inputs < 25 chars) \\
Magicoder \cite{ise_uiuc_2023} & Code & Code reasoning/generation & Generation (80\% input < 1.5k chars) \\
PAWS \cite{zhang2019pawsparaphraseadversariesword} & NLI & Paraphrase identification & 2-way (paraphrase/not), 50\% each \\
MNLI \cite{N18-1101} & NLI & Natural language inference & 3-way (0-contradiction, 1-entailment, 2-neutral), 33.3\% each \\
Flipkart Sentiment \cite{kayee_flipkart_sentiment_analysis} & Sentiment & Customer reviews & 3-way (81.2\% positive, 13.9\% negative, 4.9\% neutral) \\
Amazon Reviews \cite{xiang_zhang_acharki_yassir_2022} & Sentiment & Product reviews & 5-way (1-5 stars), 20\% each \\
IMDB Reviews \cite{maas-EtAl:2011:ACL-HLT2011} & Sentiment & Movie reviews & 2-way (1-positive, 0-negative), 50\% each \\
Pile (Toxicity) \cite{korbak2024piletoxicity} & Toxicity & Text toxicity detection & 2-way (1-toxic, 0-non-toxic), 50\% each \\
\bottomrule
\end{tabular}
\caption{Overview of datasets used in experiments.}
\label{tab:datasets_overview}
\end{table*}

\subsection{Implementation}
\label{app:implementation}
We use the Llama 3.2 3B model \cite{DBLP:journals/corr/abs-2407-21783} as our base model. Fine-tuning is performed using LoRA \cite{hu2021loralowrankadaptationlarge} with a rank of 64 and an alpha of 32, applied to $q_{proj}$, $k_{proj}$, $v_{proj}$, $o_{proj}$, $gate_{proj}$, $up_{proj}$, and $down_{proj}$ layers. We use the AdamW optimizer with a cosine learning rate schedule. The Unsloth library \cite{unsloth} is utilized for efficient training with gradient checkpointing. Experiments are tracked using Weights \& Biases (W\&B) \cite{wandb}, and vLLM \cite{kwon2023efficient} is used for optimized batch inference.

\section{LLM-as-a-Judge Prompt}
\label{sec:appendix}

The following prompt was used with Llama 3.3 70B Instruct to evaluate the model-generated outputs against the given ground truth for generation tasks. The LLM was instructed to provide a binary score (0 or 1) without explanations.

\begin{small}
\begin{verbatim}
<|begin_of_text|><|start_header_id|>
system<|end_header_id|>

Your job is to check whether the AI's answer is correct.

Compare it with the correct answer and 
score it as either 0
if the AI's answer is wrong or 1 if it is correct.

DO NOT provide any explanations.<|eot_id|>
<|start_header_id|>user<|end_header_id|>
Correct Answer: {groundtruth_column}
AI Answer: {Model generated output}<|eot_id|>

<|start_header_id|>assistant<|end_header_id|>

Score:
\end{verbatim}
\end{small}
This prompt ensured a strict, explanation-free evaluation of the model's responses based on the provided ground truth.

\section{PCA Details}
To increase the diversity of the LLM variants in the performance matrix before running PCA, we also test a few-shot baseline on Llama 3.2 1B base model. Its accuracy is 14.11 on GSM8K, 7.00 on Goat, 18.87 on Magicoder, 55.45 on PAWS, 40.35 on MNLI, 52.60 on MNLI (E), 52.55 on Pile, 90.95 on Flipkart, 48.35 on Amazon, 74.85 on IMDB. We use 5 shots for Amazon, 2 shots for Pile and IMDB, and 3 shots for the rest of the datasets. 

The rows of the linear weight matrix on the right side of \Cref{fig:pca_fig} correspond to zero-shot Llama 3.2 3B base, few-shot Llama 3.2 1B base model, LoRA from GSM8K, LoRA from Goat, LoRA from Magicoder, LoRA from PAWS, LoRA from MNLI (E), LoRA from Pile, LoRA from Flipkart, LoRA from Amazon, LoRA from IMDB (from top to bottom). Finally, before PCA, we normalize the average and standard deviation of each column to make every task, regardless of the magnitude of its accuracy, have similar importance in the PCA. 

\section{Detailed Analysis of Length Distribution Effects}
\label{app:len_dist}
\begin{figure*}[h!]
    \centering
    \begin{subfigure}[b]{0.48\textwidth}
        \includegraphics[width=\textwidth]{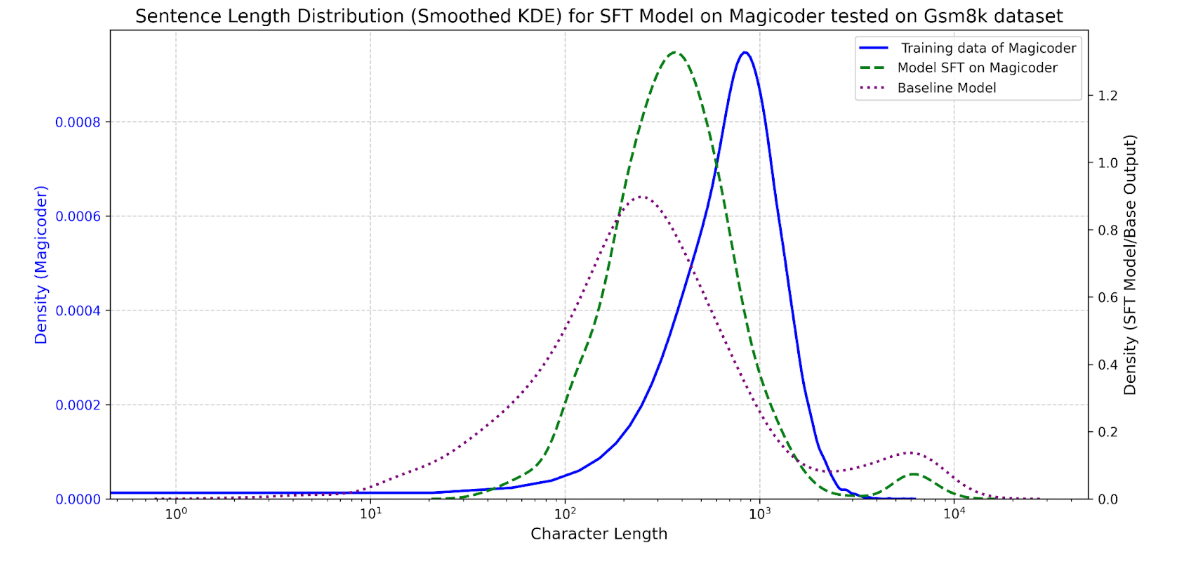} 
        \caption{\textit{M}[Magicoder], tested on GSM8K.}
        \label{fig:length_gen_task_gsm8k}
    \end{subfigure}
    \hfill
    \begin{subfigure}[b]{0.48\textwidth}
        \includegraphics[width=\textwidth]{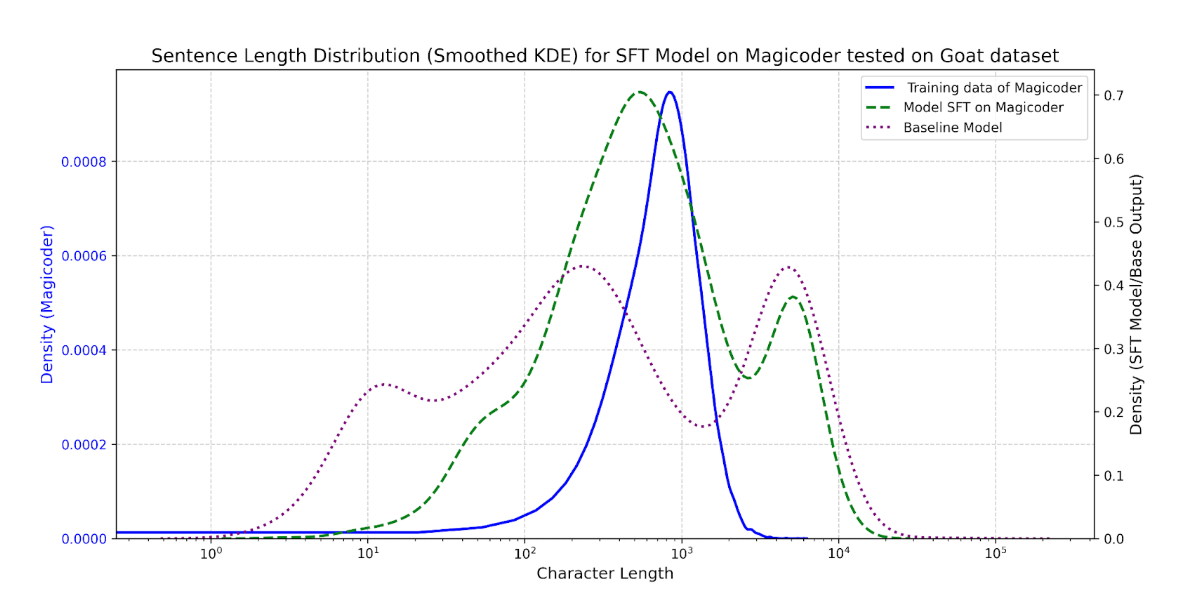}
        \caption{\textit{M}[Magicoder], tested on Goat.}
        \label{fig:length_gen_task_goat}
    \end{subfigure}
    \caption{Generation length distribution when fine-tuned on a generation task (\textit{M}[Magicoder]) and tested on other generation tasks.}
    \label{fig:length_gen_task}
\end{figure*}

\begin{figure*}[h!]
    \centering
    \begin{subfigure}[b]{0.48\textwidth}
        \includegraphics[width=1\textwidth]{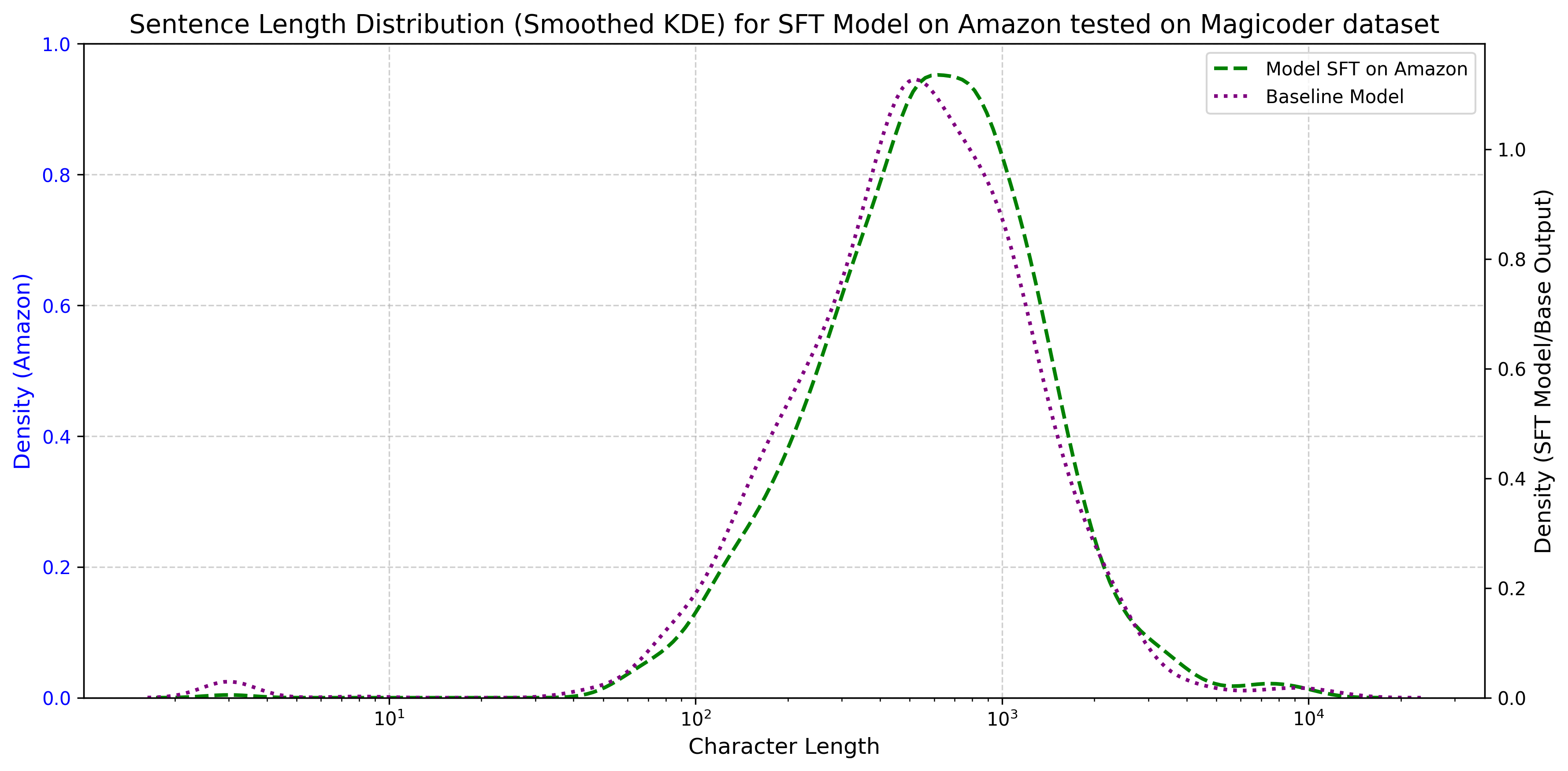} 
        \caption{\textit{M}[Amazon], tested on Magicoder.}
    \end{subfigure}
    \hfill
    \begin{subfigure}[b]{0.48\textwidth}
        \includegraphics[width=1\textwidth]{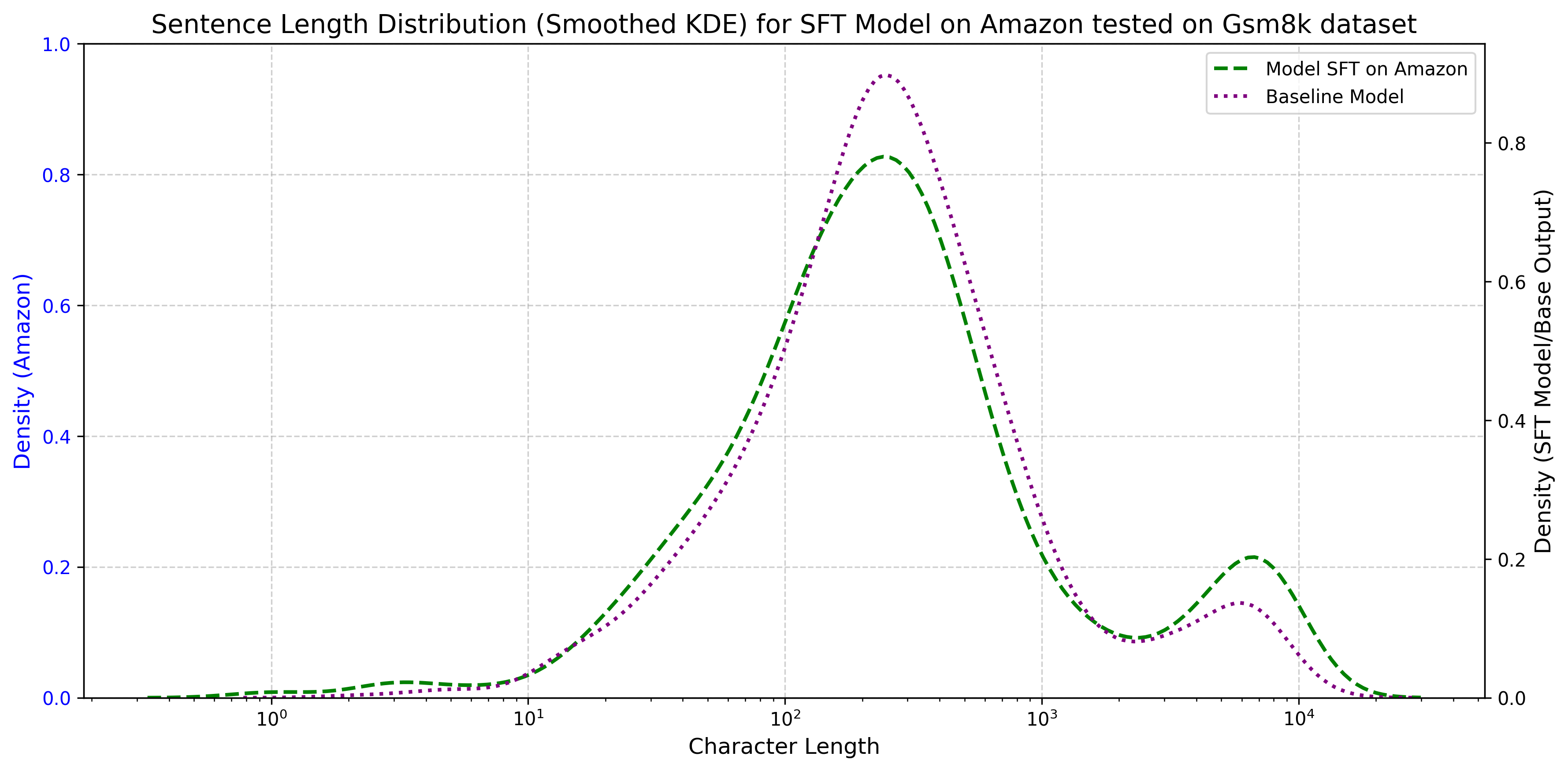} 
        \caption{\textit{M}[Amazon], tested on GSM8K.}
    \end{subfigure}
    \caption{Generation length distribution when fine-tuned on a classification task (\textit{M}[Amazon]) and tested on generation tasks.}
    \label{fig:length_class_task}
\end{figure*}

\begin{figure*}[h!]
    \centering
    \begin{subfigure}[b]{0.48\textwidth}
        \includegraphics[width=1\textwidth]{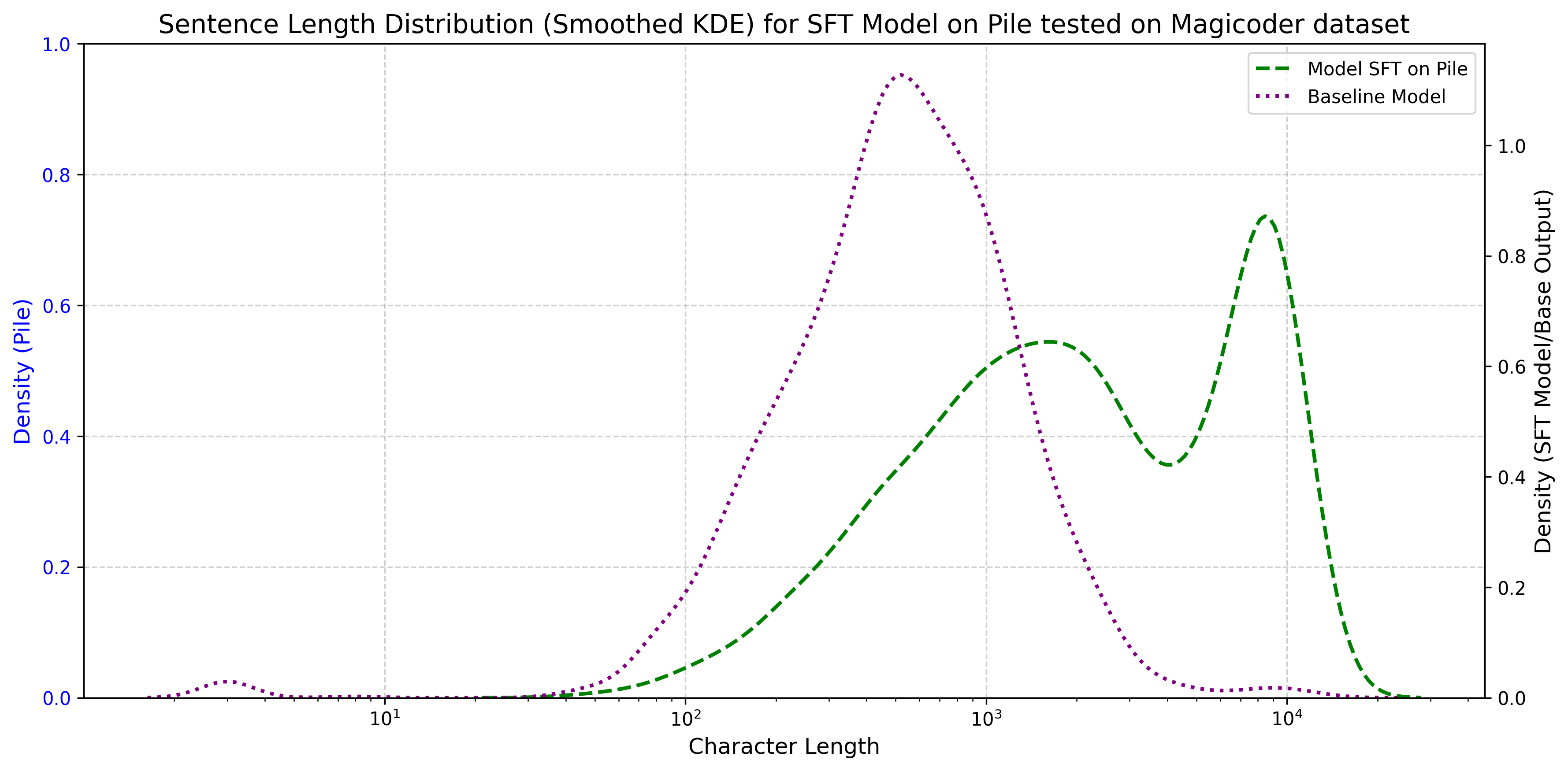} 
        \caption{\textit{M}[Pile], tested on Magicoder.}
    \end{subfigure}
    \hfill
    \begin{subfigure}[b]{0.48\textwidth}
        \includegraphics[width=1\textwidth]{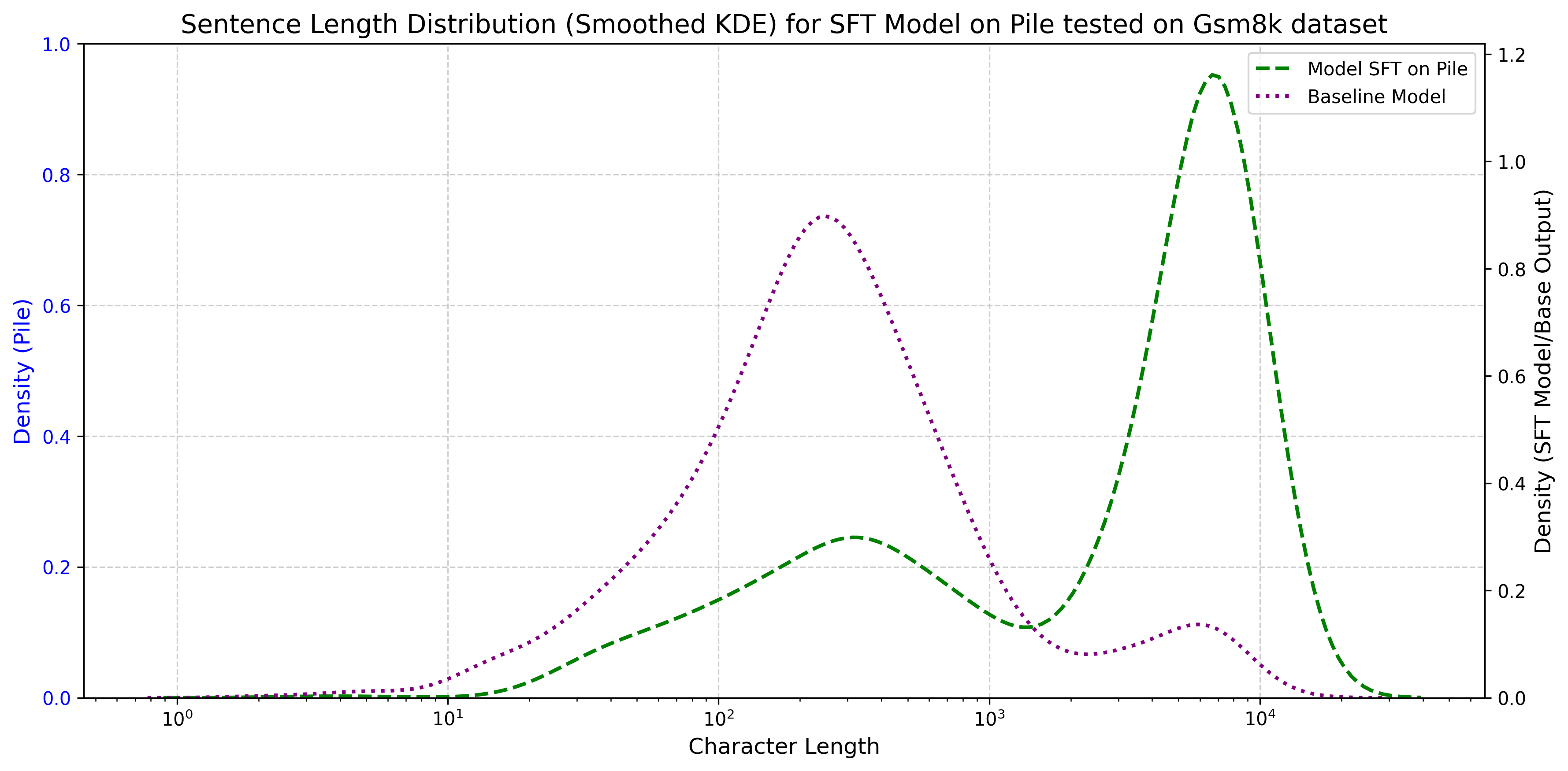} 
        \caption{\textit{M}[Pile], tested on GSM8K.}
    \end{subfigure}
    \caption{Generation length distribution when fine-tuned on Pile and tested on generation tasks.}
    \label{fig:length_pile_task}
\end{figure*}

This appendix elaborated on the observed effects of training data length distribution on model behavior, as summarized in the main paper.
\begin{itemize}
    \item \textbf{Interpolation Effect in Generation Tasks:} As shown in Figure \ref{fig:length_gen_task}, fine-tuning on generation tasks (e.g., Magicoder) leads to output lengths on other generation tasks (e.g., GSM8K, Goat) that often represent a blend. The model doesn't strictly adhere to the new dataset's length profile nor entirely retain the base model's, but rather finds an intermediate distribution.
    \item \textbf{Classification Task Influence on Generation Length:} Fine-tuning on classification tasks (e.g., Amazon Sentiment) generally preserves the base model's generation length distribution when tested on generation tasks (Figure \ref{fig:length_class_task}). The fine-tuning seems to focus more on discriminative features for classification rather than altering fundamental generative properties like typical output length, unless the classification dataset itself has a very strong and unusual length characteristic.
    \item \textbf{Dataset-Specific Tendencies:} \textit{M}[Pile] - toxicity - led to significantly longer outputs on generation tasks compared to other classification datasets, indicating that dataset-specific length characteristics can be transferred as latent traits (Figure \ref{fig:length_pile_task}).
    \item \textbf{Length Bias in Classification:} Analysis of token lengths per class (e.g., in IMDB, PAWS, Amazon - see Figure \ref{fig:category_wise_length}) reveals that base models can have biases (e.g., shorter sentences as positive sentiment). Fine-tuning can either reinforce or alter these biases depending on the training data's length characteristics per class. For example, the PAWS analysis (Figure \ref{fig:category_wise_length}) showed that the base model was biased towards shorter sentences, which Flipkart and Magicoder inherit. However, models fine-tuned on PAWS and Pile show different length biases when evaluated on IMDB (Figure \ref{fig:kde_token_lengths_imdb_comp}) Similarly, IMDB analysis (see Figure \ref{fig:category_wise_length}) showed that the base model is biased towards shorter sentences as being positive. 
\end{itemize}

\section{Length Bias for Classification Tasks}

\begin{figure*}[h!]
    \centering
    \includegraphics[width=\textwidth]
    {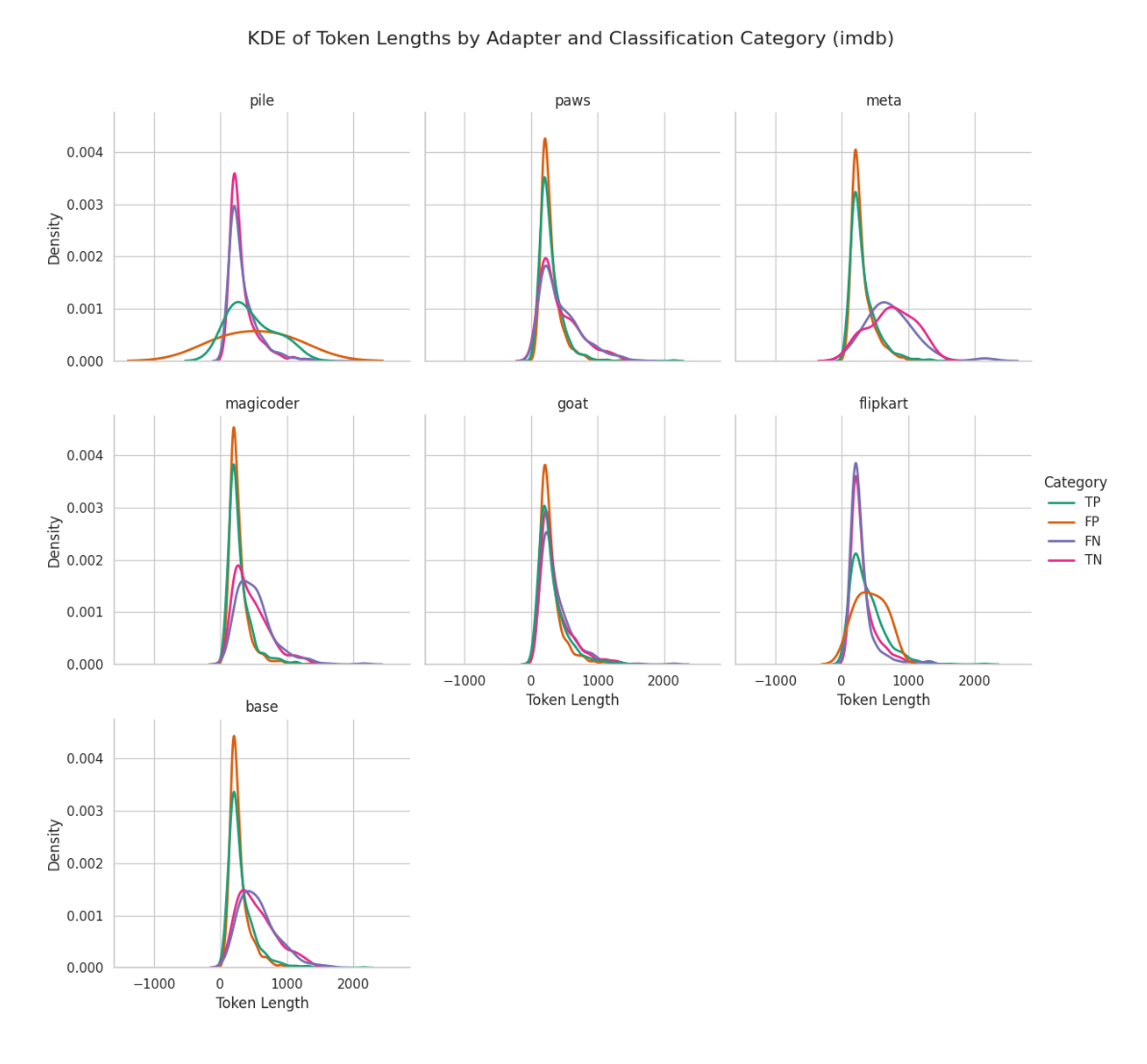}
    \caption{KDE of Token Lengths by Adapter and Classification Category (IMDB). TP: True Positive, FP: False Positive, FN: False Negative, TN: True Negative.}
    \label{fig:kde_token_lengths_imdb_comp}
\end{figure*}

\begin{figure*}[h!]
    \centering
    \includegraphics[width=\textwidth]{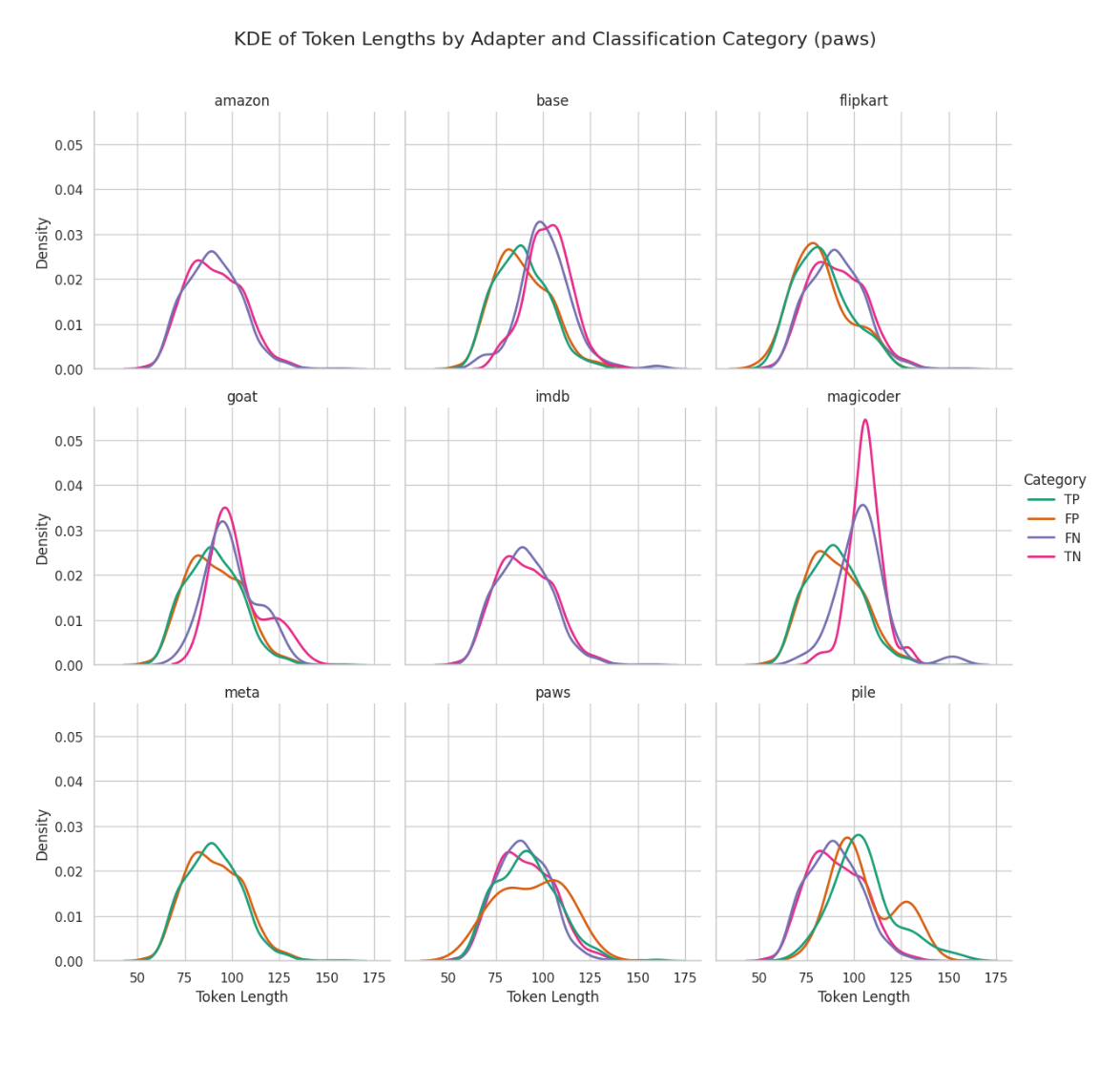}
    \caption{KDE of Token Lengths by Adapter and Classification Category (PAWS). TP: True Positive, FP: False Positive, FN: False Negative, TN: True Negative.}
    \label{fig:kde_token_lengths_paws_analysis_comp}
\end{figure*}

\begin{figure*}[h!]
    \centering
    \begin{subfigure}[b]{0.48\textwidth}
        \includegraphics[width=1\textwidth]{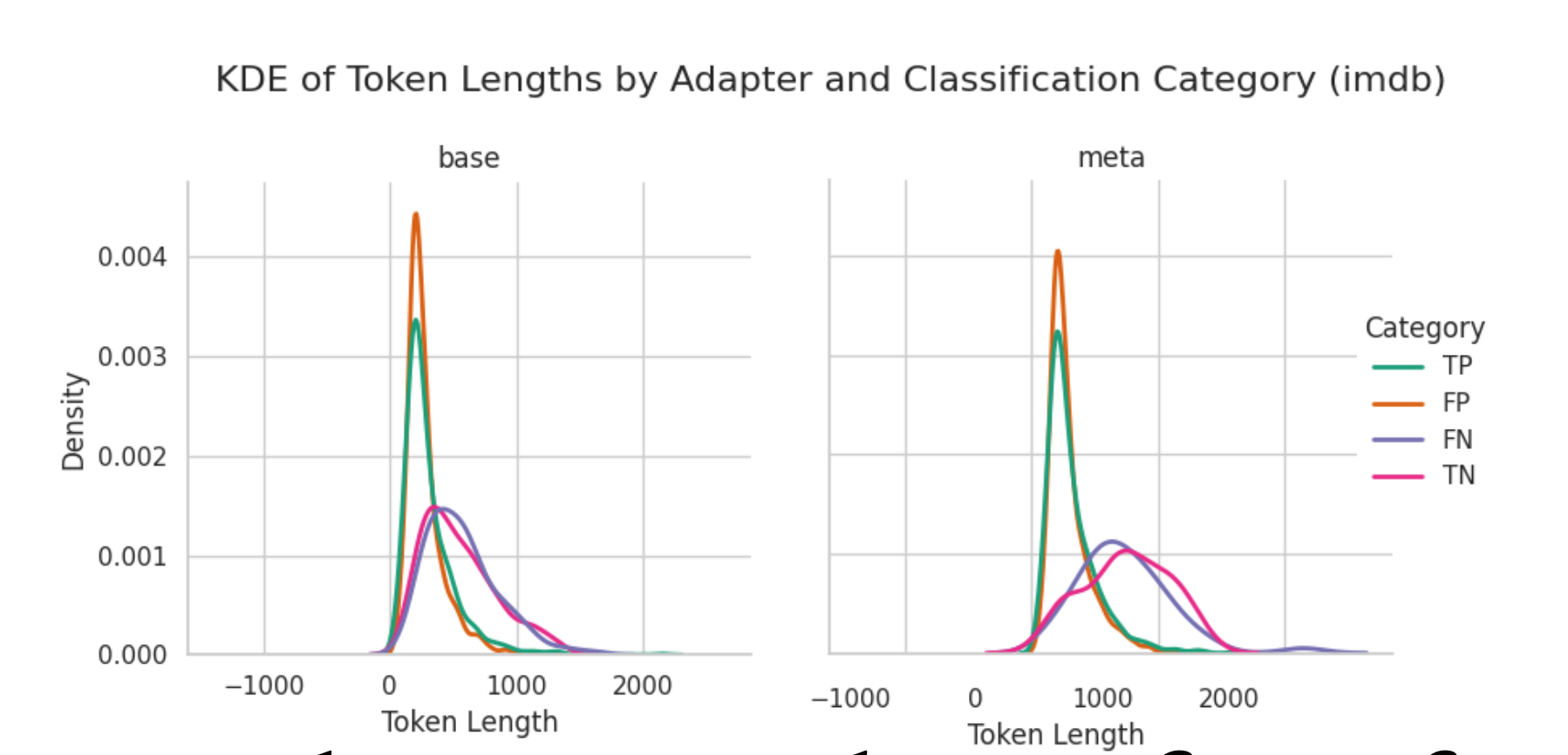} 
        \caption{KDE of Token Lengths of Original LLM and Meta-Math on IMDB}
    \end{subfigure}
    \hfill
    \begin{subfigure}[b]{0.48\textwidth}
        \includegraphics[width=1\textwidth]{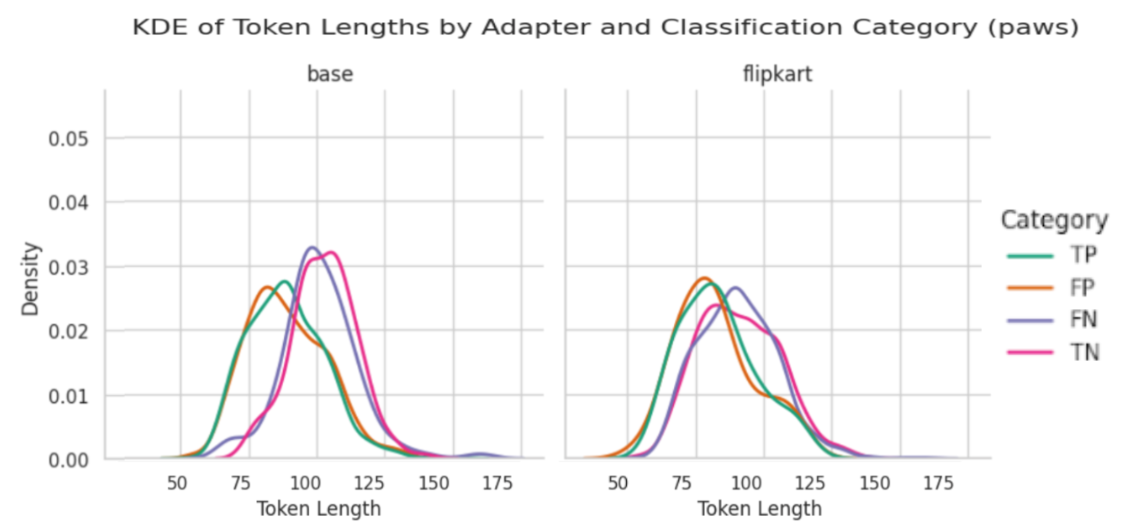} 
        \caption{KDE of Token Lengths of Original LLM and Flipkart on IMDB}
    \end{subfigure}
    \caption{KDE of Token Lengths by Adapter and Classification Category. TP: True Positive, FP: False Positive, FN: False Negative, TN: True Negative.}
    \label{fig:category_wise_length}
\end{figure*}

Figures \ref{fig:kde_token_lengths_imdb_comp}, \ref{fig:kde_token_lengths_paws_analysis_comp}, \ref{fig:category_wise_length} show how the base and adapter do in predicting classes in a classification task and their distribution across True Positive, False Positive, True Negative and False Negative. 

\section{Detailed Analysis of Classification Adapter Effects on Math Performance}
\label{app:classification_math_details}

This appendix provides a more detailed look at how fine-tuning on classification datasets impacts performance on mathematical reasoning tasks, particularly GSM8K.

\subsection{Linguistic Feature Importance Details}
\begin{figure}[h!]
    \centering
    \includegraphics[width=\columnwidth]{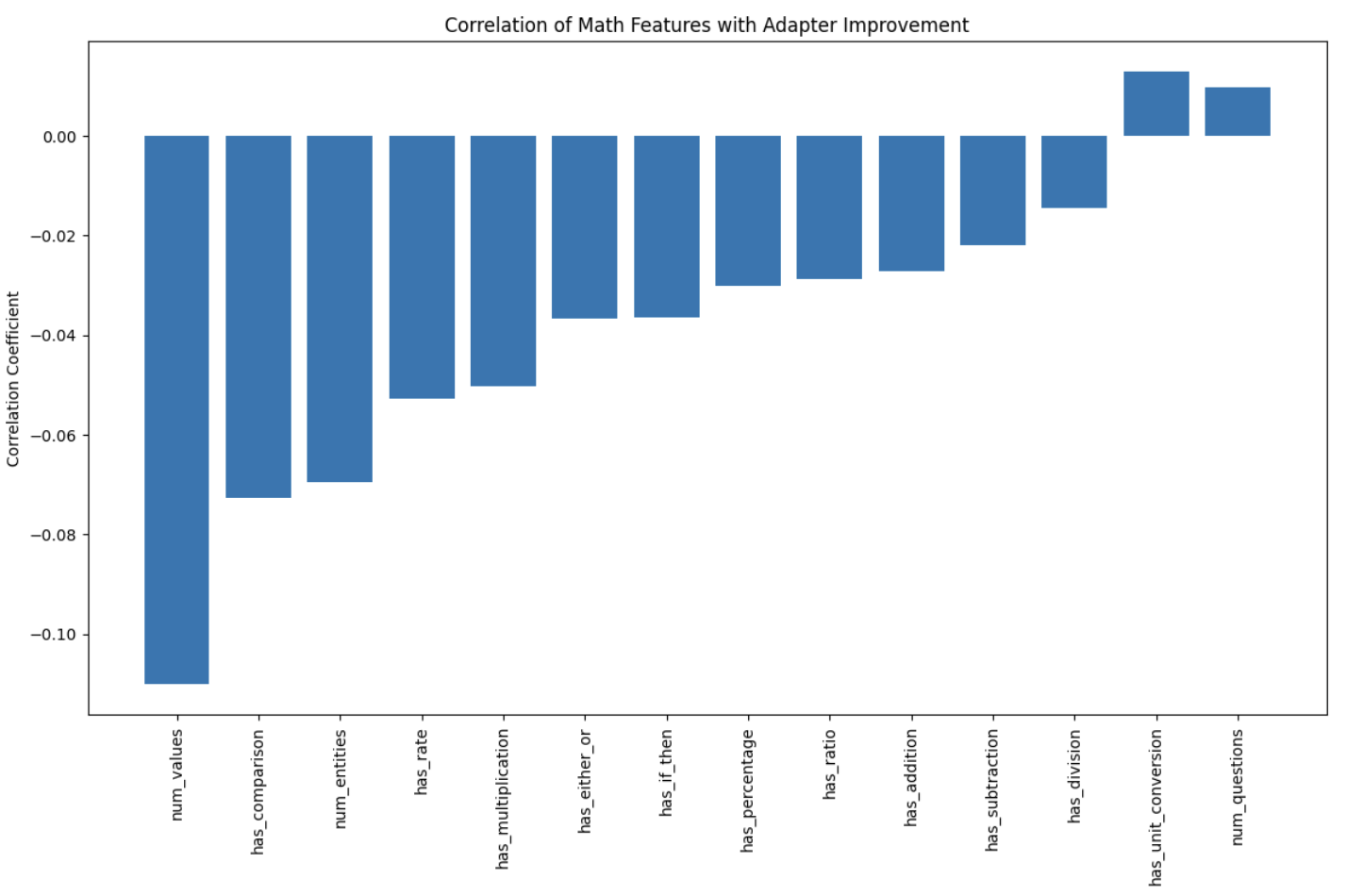} 
    \caption{Correlation of Math Features in Word Problems with Adapter Improvement on GSM8K.}
    \label{fig:corr_math_features_appendix}
\end{figure}

\begin{figure}[h!]
    \centering
    \includegraphics[width=\columnwidth]{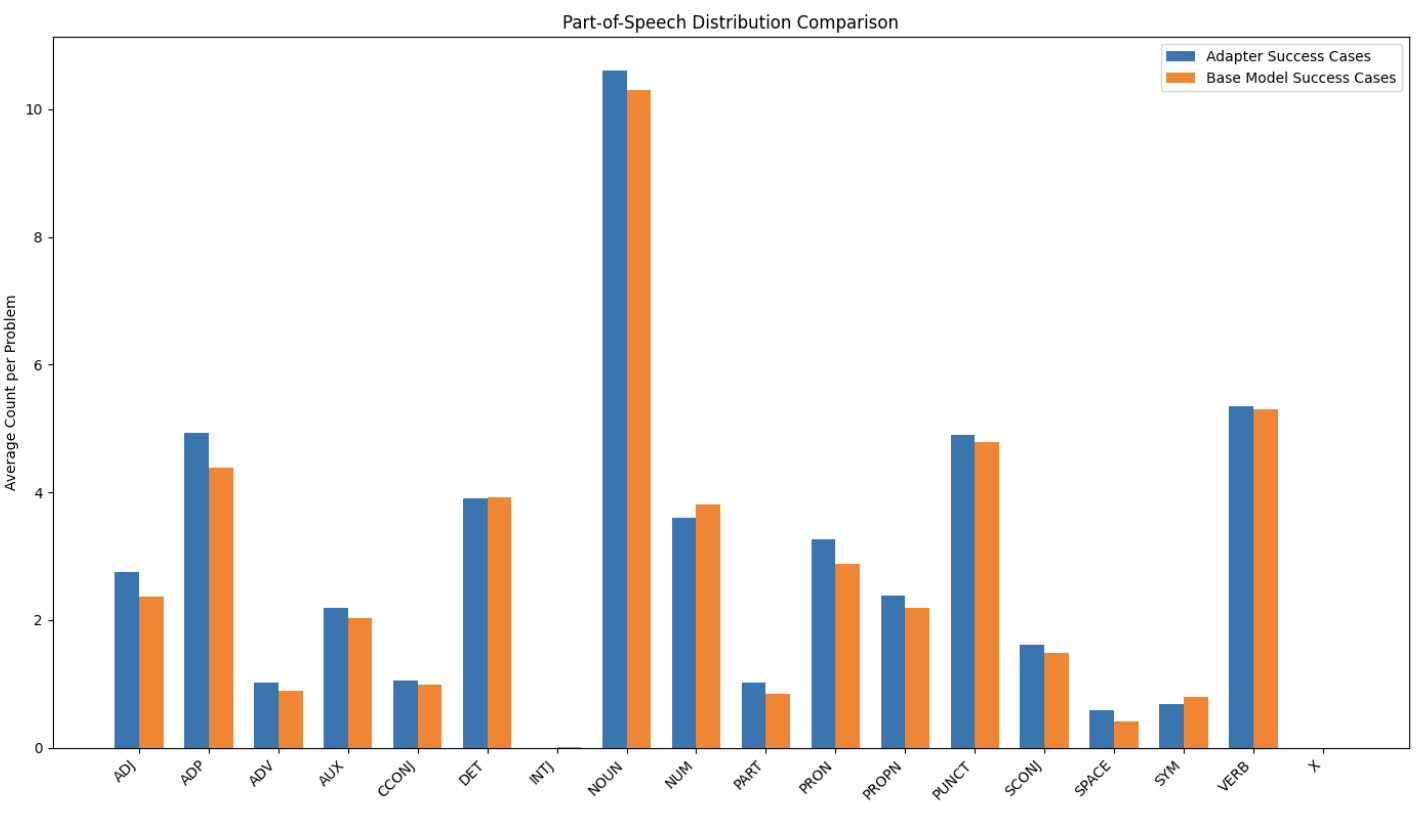} 
    \caption{Part-of-Speech Distribution Comparison in GSM8K Problems (Adapter Success Cases vs. Base Model Success Cases).}
    \label{fig:pos_distribution_appendix}
\end{figure}

The improvement from classification adapters on GSM8K appears linked to enhanced sensitivity to linguistic structures crucial for understanding word problems.
\begin{itemize}
    \item \textbf{Correlation of Math Features:} We analyzed the correlation between various mathematical features in word problems 
    and the improvement in model performance when using adapters (Figure \ref{fig:corr_math_features_appendix}). Features like "num\_values", "has\_comparison", and "num\_entities" showed negative correlations, suggesting problems with these features are less likely to show improvement with the tested adapters. Conversely, features like "has\_unit\_conversion" and "num\_questions" showed positive (or less negative) correlations, indicating adapters might handle these better.

    \item \textbf{Part-of-Speech (POS) Tags:} Comparing POS tag distributions in problems solved successfully by adapter-tuned models versus the base model reveals differences (Figure \ref{fig:pos_distribution_appendix}). For instance, if adapter success cases show a higher count of \textit{NOUNs}, it suggests adapters better handle noun-rich problems.
\end{itemize}

\begin{figure}[h!]
    \centering
    \includegraphics[width=\columnwidth]{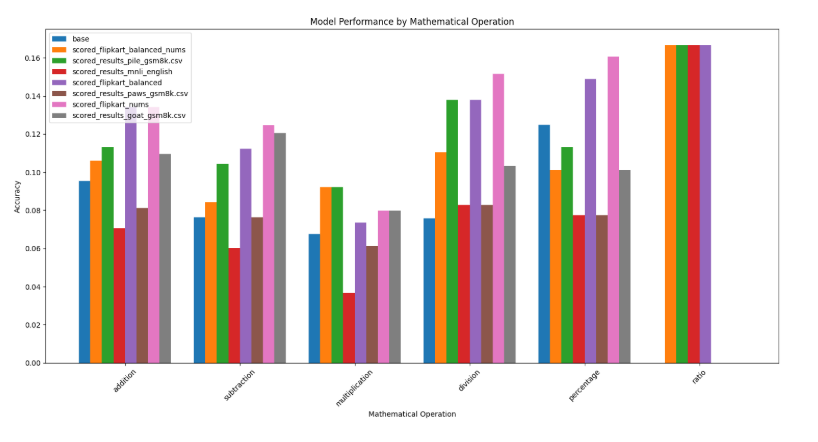}
    \caption{Adapter wise performance improvement on GSM8K clustered by arithmetic operations.}
    \label{fig:math_cluster}
\end{figure}

\subsection{Analysis by Number of Steps}
\label{app:steps_appendix} 
\begin{figure}[h!]
    \centering
    \includegraphics[width=\columnwidth]{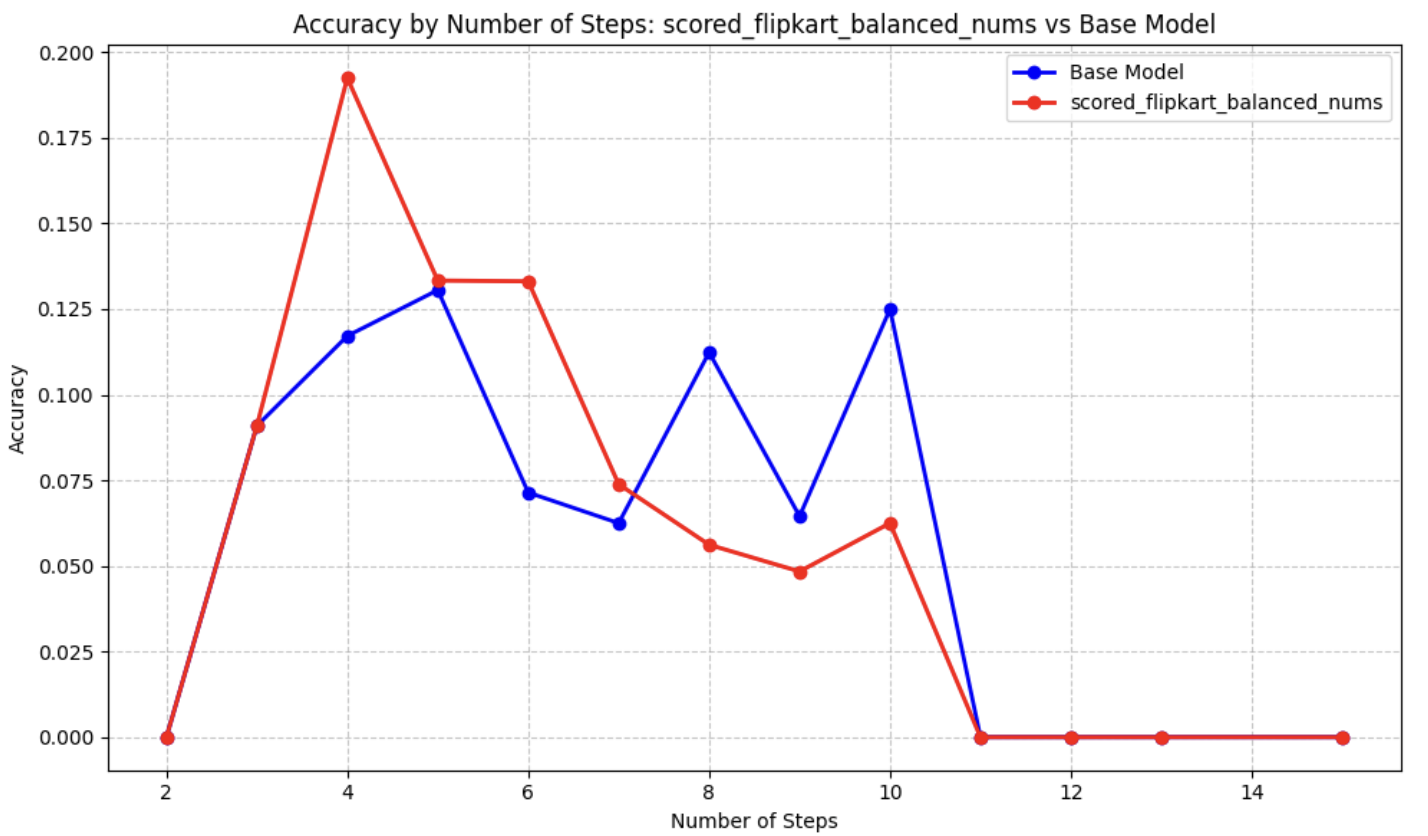} 
    \caption{Accuracy by Number of Steps: M[Flipkart Balanced (Numeric)] vs Original LLM on GSM8K.}
    \label{fig:accuracy_by_steps_example_appendix}
\end{figure}

\begin{figure}[h!]
    \centering
    \includegraphics[width=\columnwidth]{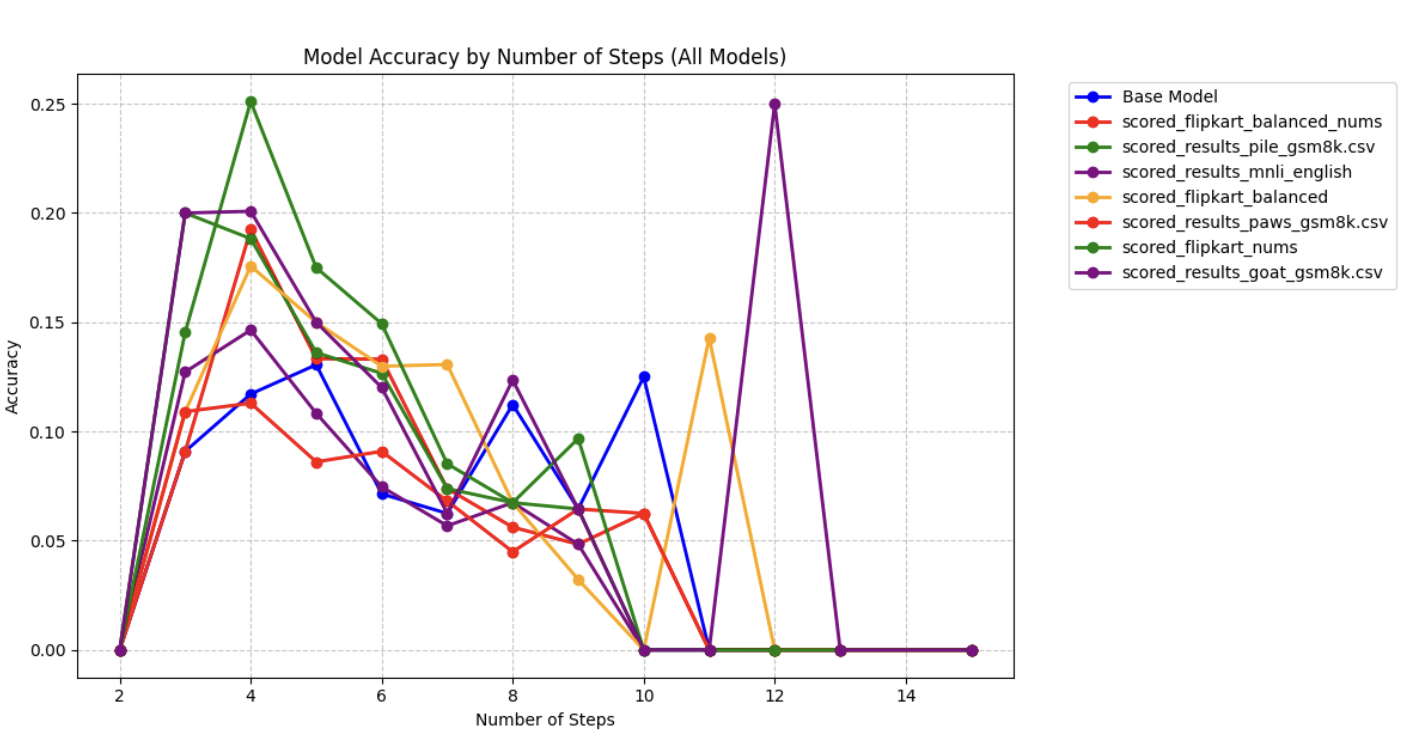} 
    \caption{Model Accuracy by Number of Steps on GSM8K, highlighting the 10-step region.}
    \label{fig:accuracy_by_steps_10_detail_appendix}
\end{figure}

\begin{figure*}[h!]
    \centering
    \begin{subfigure}[b]{0.48\textwidth}
        \includegraphics[width=\columnwidth]{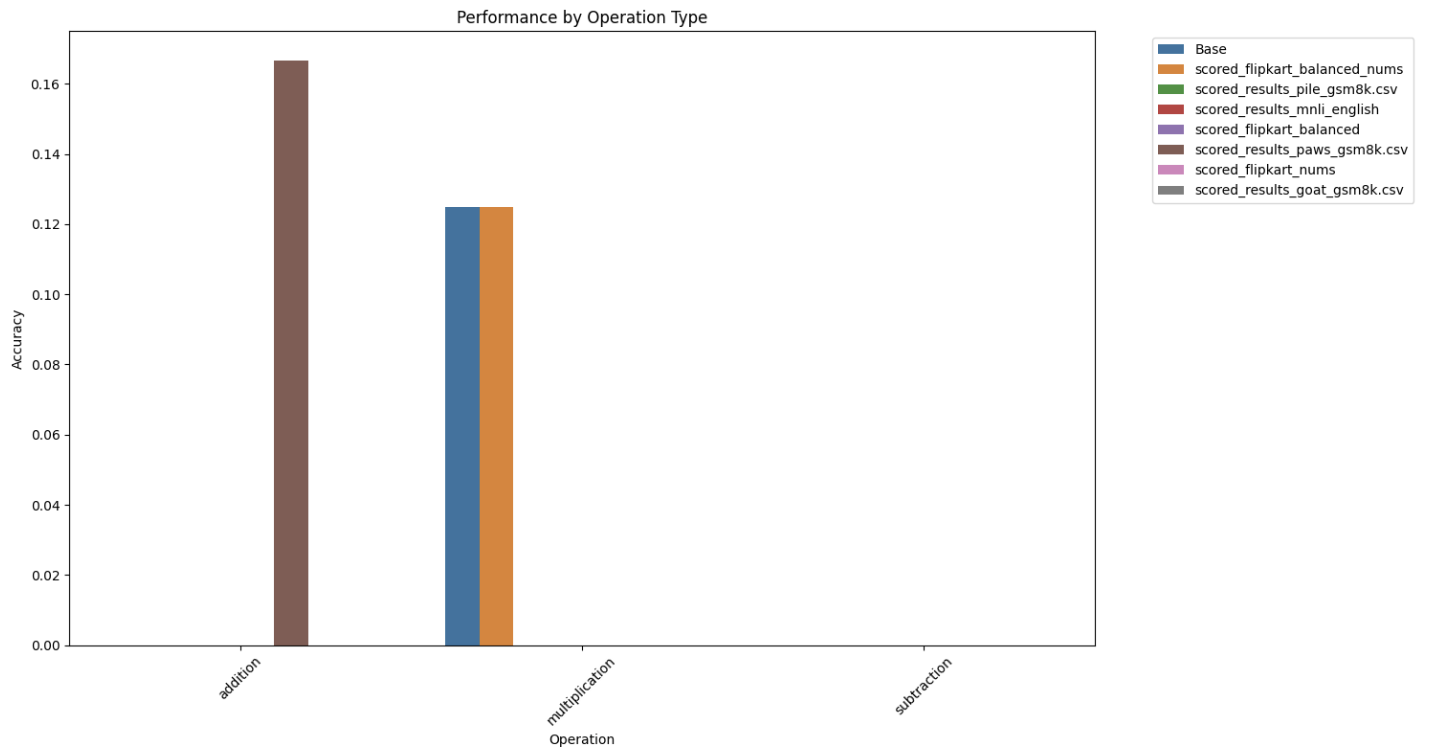} 
        \caption{Performance by Operation Type for 10-Step Problems.}
        \label{fig:perf_op_type_10_steps_appendix}
    \end{subfigure}
    \hfill
    \begin{subfigure}[b]{0.48\textwidth}
        \includegraphics[width=\columnwidth]{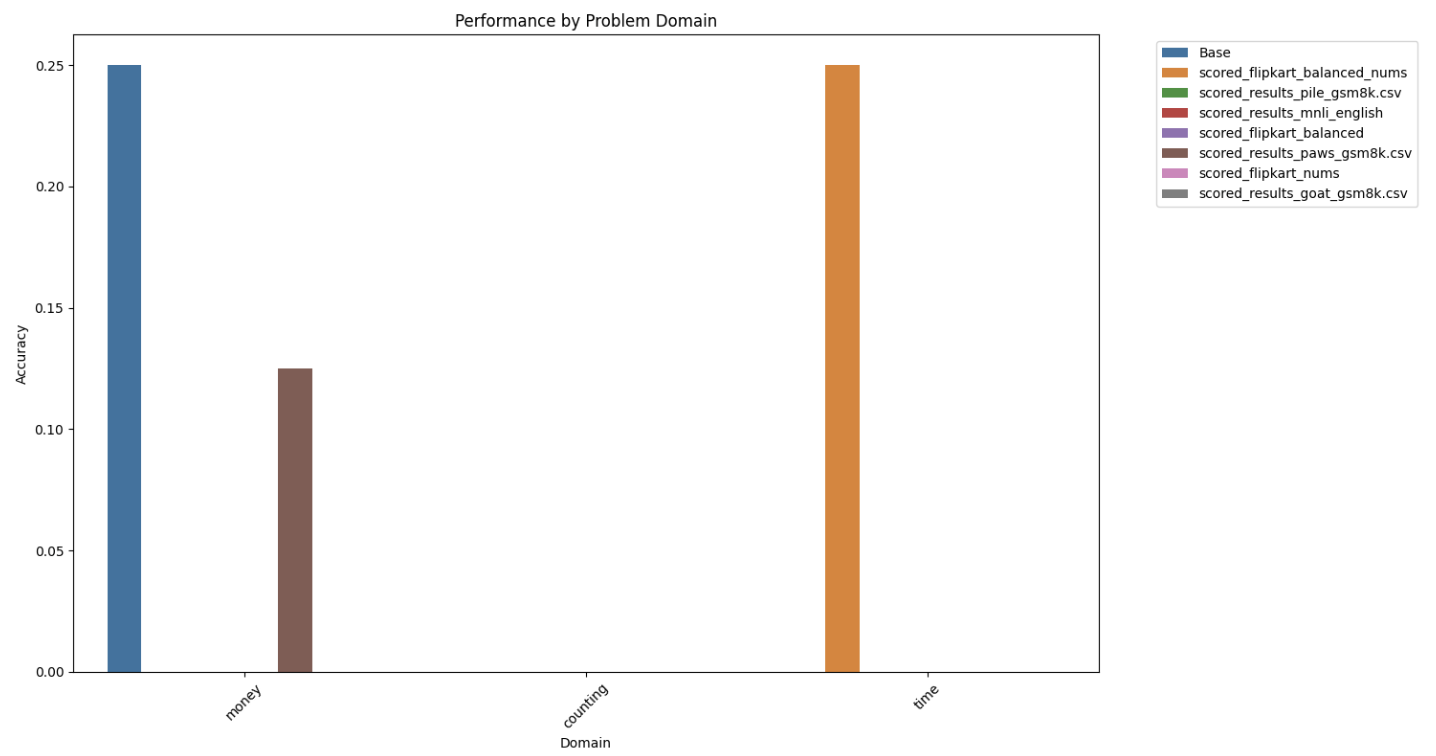} 
        \caption{Performance by Problem Domain for 10-Step Problems.}
        \label{fig:perf_domain_10_steps_appendix}
    \end{subfigure}
    \caption{Analysis of 10-Step GSM8K problems where base model outperforms adapters.}
    \label{fig:analysis_10_steps_appendix}
\end{figure*}

We defined a heuristic "number of steps" to loosely quantify problem complexity by summing counts of questions, explicit sentences, mathematical operations, comparisons, and conditional statements.
\begin{itemize}
    \item Generally, adapter-tuned models showed varying performance improvements over the base model depending on the number of steps, often outperforming for lower to moderate step counts (Figure \ref{fig:accuracy_by_steps_example_appendix}).
    \item A peculiar dip in adapter performance relative to the base model was consistently observed for problems estimated to have 10 steps (Figure \ref{fig:accuracy_by_steps_10_detail_appendix}). Analysis of these 10-step problems revealed they predominantly involved `money' domain and `multiplication' or `addition' operations (Figures \ref{fig:perf_op_type_10_steps_appendix}, \ref{fig:perf_domain_10_steps_appendix}). The base model excelled on these specific 10-step problems, while adapter performance decreased. 
\end{itemize}

\end{document}